\documentclass[10pt,twocolumn,letterpaper]{article}

\usepackage{cvpr}
\usepackage{multicol}
\usepackage{multirow}
\usepackage{xspace}
\usepackage{amsmath}
\usepackage{amssymb}
\usepackage{colortbl}
\definecolor{mygray}{gray}{.9}
\usepackage{color}

\usepackage{listings} 
\usepackage{xcolor}
\definecolor{lightgray}{rgb}{0.95, 0.95, 0.95}
\usepackage{tcolorbox}
\usepackage{ragged2e}
\definecolor{boxheader}{HTML}{1A2B5E}
\definecolor{boxborder}{HTML}{2854A3}
\definecolor{boxbackground}{HTML}{F8F8F8}
\newtcolorbox{promptbox}[1]{
    colback=boxbackground,
    colframe=boxborder,
    coltitle=white,
    boxrule=1pt,
    arc=4mm,
    outer arc=4mm,
    colbacktitle=boxheader,
    fonttitle=\bfseries\small\sffamily,
    left=3mm, right=3mm, top=2mm, bottom=2mm,
    title={#1},
    parbox=false,
}

\definecolor{cvprblue}{rgb}{0.21,0.49,0.74}
\usepackage[pagebackref,breaklinks,colorlinks,allcolors=cvprblue]{hyperref}
\usepackage{algpseudocode}

\usepackage{amsmath}
\usepackage{graphicx}
\usepackage{float}
\usepackage{multirow}
\usepackage{listings}
\usepackage{comment}

\def\paperID{3900}
\def\confName{CVPR}
\def\confYear{2026}

\title{Thinking by Doing: Building Efficient World Model Reasoning in LLMs \\ via Multi-turn Interaction}

\author{
Bao Shu$^{1*\ddagger}$, 
Yan Cai$^{2*\ddagger}$, 
Jianjian Sun$^3$, 
Chunrui Han$^3$, 
En Yu$^{3}$, \\
Liang Zhao$^3$,
Jingcheng Hu$^4\ddagger$, 
Yinmin Zhang$^3$, 
Haoran Lv$^3$,
Yuang Peng$^{3}$, \\
Zheng Ge$^3$, 
Xiangyu Zhang$^3$, 
Daxin Jiang$^3$, 
Xiangyu Yue$^{1\dagger}$\\
$^1$CUHK MMLab, $^2$Peking University, $^3$StepFun, $^4$Tsinghua University.
}

\newcommand{\methodfull}{\textbf{w}orld-\textbf{m}odel internalization through efficient interaction and \textbf{act}ive reasoning}
\newcommand{\methodabb}{WMAct}

\begin{document}

\maketitle
\let\thefootnote\relax
\footnotetext{$^{*}$Equal contribution.}
\footnotetext{$^{\dagger}$Corresponding author.}
\footnotetext{$^{\ddagger}$Work performed during an internship at StepFun.}

\begin{abstract}

Developing robust world model reasoning is crucial for large language model (LLM) agents to plan and interact in complex environments. 
While multi-turn interaction offers a superior understanding of environmental dynamics via authentic feedback, current approaches often impose a rigid reasoning process, which constrains the model's active learning, ultimately hindering efficient world model reasoning.
To address these issues, we explore \methodfull~(\methodabb), which liberates the model from structured reasoning—allowing the model to shape thinking directly through its doing—and achieves effective and efficient world model reasoning with two key mechanisms:
(1) a reward rescaling mechanism adjusting outcome reward based on action efficacy to incentivize redundancy reduction and purposeful interaction; 
(2) an interaction frequency annealing strategy to progressively reduce the maximum allowed interaction turns, which compels the model to condense its learning and internalize environmental dynamics rather than over-relying on environmental cues. 
Our experiments on Sokoban, Maze, and Taxi show that \methodabb~yields effective world model reasoning capable of resolving tasks in a single turn that previously required multiple interactions and fosters strong transferability to complex environments, improving performance on a suite of reasoning benchmarks.
\end{abstract}
    
\section{Introduction}
\label{sec:intro}
Recent advancements have been marked by the development of sophisticated reasoning large language models (LLMs), such as OpenAI-o1~\cite{jaech2024openai} and Deepseek-R1~\cite{guo2025deepseek}, which are distinguished by a protracted reasoning process before delivering a final answer. This reasoning capability, largely unlocked by Reinforcement Learning (RL), has sparked broad research enthusiasm and driven substantial advances in demanding domains, such as mathematics~\cite{zeng2025simplerlzooinvestigatingtamingzero,hu2025open,yu2025dapo,skywork-or1-2025, wei2024slow}, coding~\cite{pourreza2025reasoningsql,wei2025swe,qwen3technicalreport}, logic~\cite{xie2025logic,sprague2024cot,cheng2025revisiting, wei2025open}, and multimodal~\cite{yu2025perceptionr, yu2025unhackable, zhu2025perpo}.

Building upon these sophisticated reasoning capabilities, the research focus is increasingly shifting to empower LLMs as autonomous agents, thereby introducing the critical challenge of agentic tasks.
In these tasks, models must not only deduce solutions but also formulate sequential plans, generate executable actions, and dynamically interact, which challenges their grasp of world knowledge, such as environmental properties and state transitions. 
Consequently, much recent effort has focused on designing explicitly structured world model reasoning processes, such as task planning~\cite{zhang2025rlvmr}, environment perception~\cite{chen2025g1, wang2025vagen, zhang2025rlvmr,liao2025think}, and state change prediction~\cite{wang2025vagen}
, and employing RL to reinforce these patterns, thereby enhancing the model's understanding of environmental dynamics. 
While these predefined, human-designed cognitive patterns may prove effective in the short term, they ultimately become a bottleneck, limiting the model's ability to spontaneously learn strategies from its interactions with the environment~\cite{silver2017mastering,Silver2017MasteringTG,Sutton2019BitterLesson}.
To this end, we investigate a key question: \textbf{How can we build effective and efficient world model reasoning without constraining the model's cognitive flexibility?}
\begin{figure*}
    \centering
    \includegraphics[width=\linewidth]{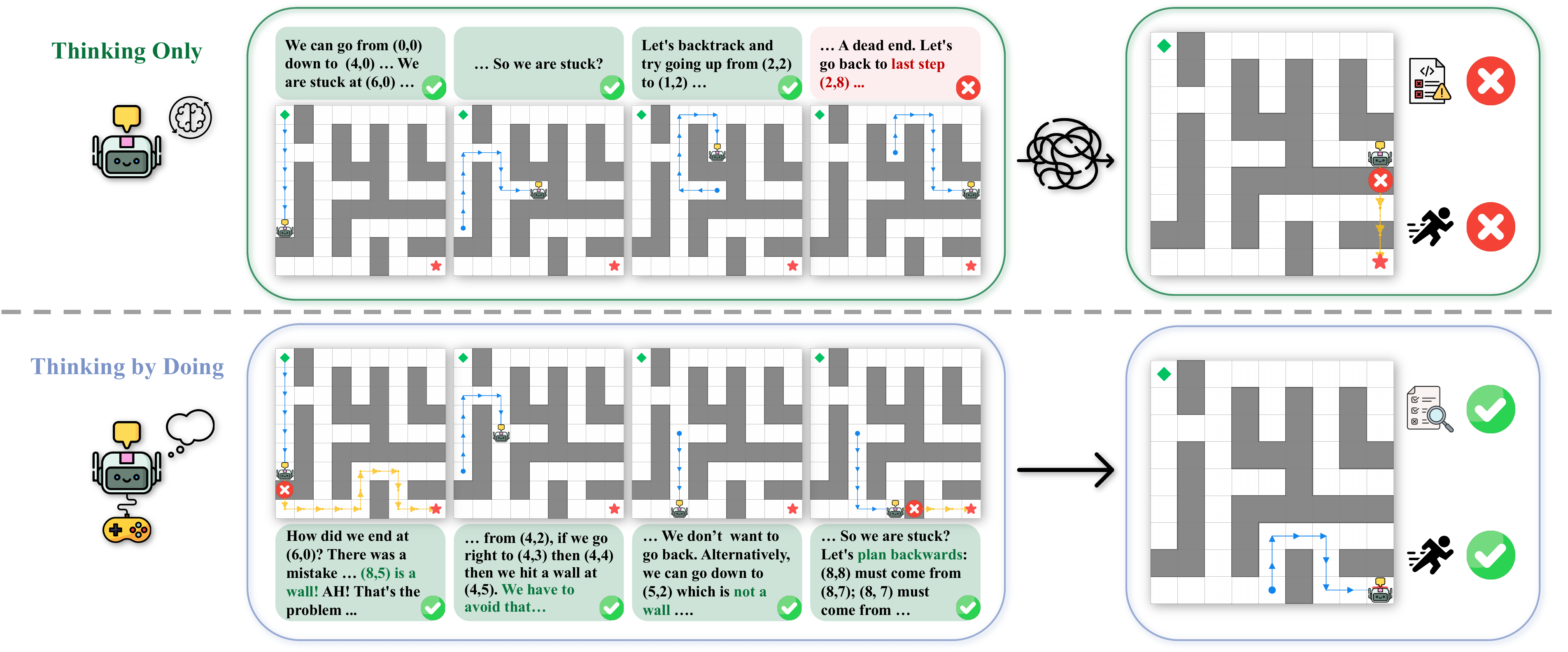}
    \caption{Illustration about the difference between monolithic reasoning (Thinking Only) and multi-turn interaction approach (Thinking by Doing). Top: the agent relies on monolithic reasoning and internal simulation to plan a path. This strategy imposes a substantial cognitive burden without interaction, risking the reinforcement of erroneous internal knowledge and ultimately leading to failure. Bottom: Multi-turn interaction avoids pitfalls of internal simulation, allowing for continuous path validation and correction, resulting in successful completion.}
    \label{fig:singleverusmulti}
    \vspace{-1em}
\end{figure*}

Our investigation to elucidate this issue involved two RL training paradigms: monolithic reasoning, which requires the model to generate an entire plan for a given task, and multi-turn interaction, wherein the model obtains intermediate feedback through interaction with the environment to actively respond to state transition. Through this analysis, we identified the following three challenges:
(1) \textit{Substantial cognitive burden without interaction.} Monolithic reasoning, while employing sophisticated strategies like hypothesis generation and validation, suffers from a substantial cognitive burden stemming from the necessity of internally simulating continuous state transitions. This internal simulation without external interaction introduces significant risks of reinforcing self-generated and erroneous environmental knowledge, culminating in spurious success. In this scenario, the model may exhibit strong performance in the training environment but ultimately fails to generalize to unseen scenarios. However, the process of multi-turn interaction, coupled with advanced reasoning patterns, enables the model to leverage environmental feedback for self-evolving and subsequently internalize it into the world model reasoning process. 
(2) \textit{Inefficient exploration strategies leading to redundant world knowledge.} Models adopt inefficient brute-force strategies to interact with the environment, enumerating numerous actions rather than analyzing the optimal path. These meaningless operations impede the model's ability to acquire higher-quality environmental knowledge from the feedback provided by the environment in response to its actions, thereby undermining more efficient world model reasoning.
(3) \textit{Excessive interaction dependency causes inefficient internalization of world model reasoning}. LLMs tend to exploit feedback obtained from interaction solely for task completion, lacking the incentive to internalize more complex environmental dynamics into their own reasoning processes, thus hindering the ability to perform efficient long-horizon world model reasoning.

\begin{figure*}
    \centering
    \includegraphics[width=\linewidth]{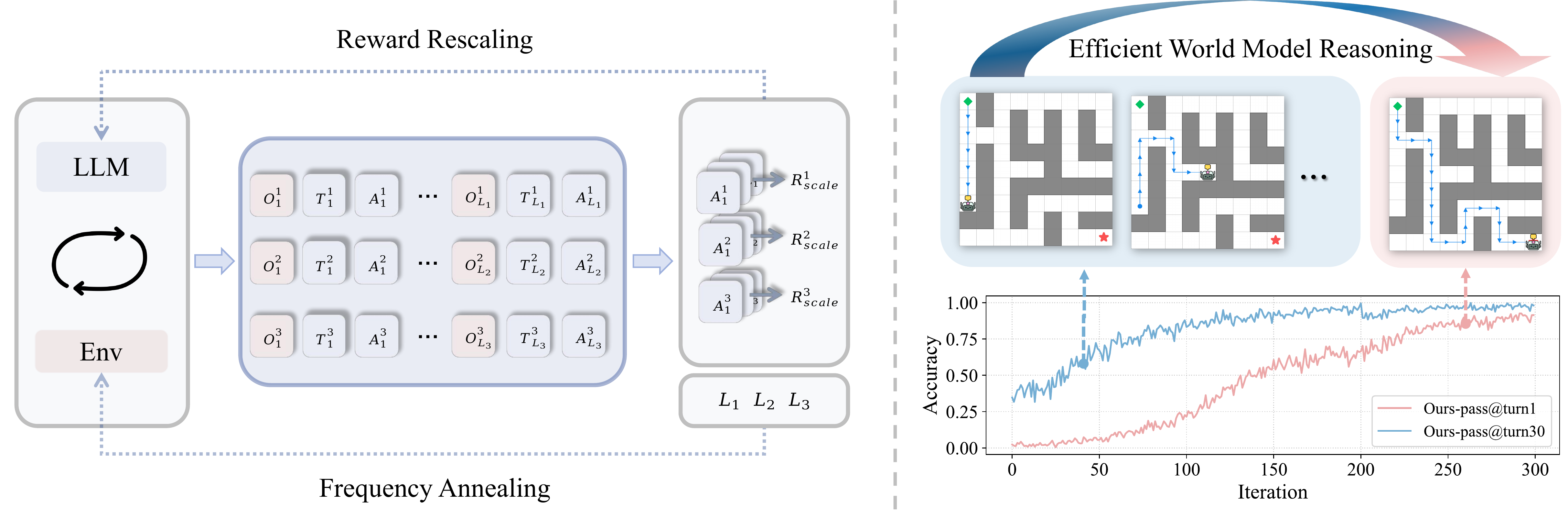}
    \caption{
Illustration about the evolution of the model's behavior in the Maze example. As training progresses, the model's single-turn accuracy continually improves, ultimately matching multi-turn accuracy. Tasks that previously necessitated multi-turn interactive trial-and-error can now be effectively solved within a single turn.
This progression illustrates a transition from multi-turn, reactive to single-turn, proactive planning. The model first relies on step-by-step interaction but later enhances its long-range planning capabilities, allowing it to internalize the complex interactive strategy, thereby improving efficiency and computational economy without sacrificing accuracy.
}
    \label{fig:efficient}
    \vspace{-1em}
\end{figure*}

In this work, we formally present \methodfull~(\methodabb), which helps models learn world model reasoning from multi-turn interaction and addresses the aforementioned problems through two key mechanisms. 
To incentivize the agent to reduce redundant actions and learn purposeful interactive strategies, we introduce a reward rescaling mechanism. This mechanism scales the outcome reward by the proportion of effective actions, forcing the agent to prioritize concise strategies. 
In addition, we employ an interaction frequency annealing strategy, progressively adjusting the allowed interaction turns, thereby encouraging initial environmental exploration while later compelling the model to distill and internalize environmental dynamics.
The synergy of these two mechanisms yields a more efficient world model reasoning process capable of resolving tasks in a single turn that previously required multiple interactions demonstrated in Figure~\ref{fig:efficient}, as validated in the Sokoban, Maze, and Taxi environments.

\section{Related Work}
\label{sec:related_work}
\subsection{World Model Reasoning}
Empowering LLMs as autonomous agents requires world knowledge such as environment perception and state transition prediction.
Approaches to building world models in LLMs are diverse.
One line of work injects knowledge through auxiliary tasks during pre-training or fine-tuning.
For instance, Agentic CPT~\cite{su2025scaling} builds an entity-anchored knowledge memory from diverse data sources to synthesize multi-style question-answers.
ScaleCUA~\cite{liu2025scalecua} introduces a suite of GUI-related tasks to comprehensively lay a foundation for the digital world.
The early experience paradigm~\cite{zhang2025agent} constructs next-state prediction samples from agent trajectories.
A different line of work attempts to improve reasoning by enforcing a predefined, structured reasoning process during interaction.
These methods compel the model to follow a fixed cognitive pattern designed by humans.
G1~\cite{chen2025g1} forces the model to first output perceptual information, providing a reliable state foundation for subsequent structured reasoning and achieving the mutual bootstrapping of perception and reasoning abilities.
RLVMR~\cite{zhang2025rlvmr} provides dense process-level rewards for explicit structured meta-reasoning steps, effectively enhancing the error recovery capabilities and generalization to unseen tasks.
VAGEN~\cite{wang2025ragen} explicitly forces agents to generate StateEstimation and TransitionModeling, and reinforces this process using a specialized world-modeling reward, leading to superior comprehension of environments.
Although structured reasoning works, these rigid patterns prevent models from learning flexible, efficient strategies through interaction. Our research focuses on enabling models to develop reasoning capabilities more autonomously.

\subsection{Multi-turn reinforcement learning}
The rapid advancement of LLMs has facilitated the creation of autonomous agents capable of performing complex tasks~\cite{team2025kimi,guo2025seed1, wang2025internvl3,hong2025glm,wang2025step,qwen3technicalreport}. These agents typically operate through multi-turn interactions, requiring dynamic reasoning, action execution, and adaptation based on environmental feedback. This paradigm has motivated the development of multi-turn reinforcement learning algorithms aimed at training such agents effectively. For example, Zhou et al. \cite{zhou2024archer} propose ArCHer, a hierarchical RL method that improves sample efficiency by separating high-level and low-level policies. Wang et al. \cite{wang2025ragen} introduce StarPO, which optimizes at the trajectory level and mitigates a common failure mode called echo trap through variance-based filtering. Xue et al. \cite{xue2025simpletir} attribute instability in tool-integrated reasoning to distributional drift from tool feedback, leading to void turns and stabilizing training by excluding affected trajectories. Similarly, Shang et al. \cite{shang2025rstar2} present GRPO-RoC, a strategy that resamples correct trajectories to reduce noise from code execution environments during policy updates.
While these methods focus on improving the stability and sample efficiency of the RL algorithm itself for multi-step task completion, our primary goal is not merely to complete the task but to compel the model to internalize environmental dynamics, ultimately improving long-horizon capability.

\section{Method}
\label{sec:method}
\subsection{Background}
RL provides a formal framework for an agent to learn an optimal policy $\pi_{\theta}$ through environmental interaction, with the objective of maximizing the expected cumulative reward.
For language models, a trajectory typically involves generating a response $y$ from a given prompt $x$. 
The policy gradient theorem~\cite{NIPS1999_464d828b,712192} offers a way to optimize this objective by directly computing the gradient of $J(\theta)$ with respect to the policy parameters $\theta$. The gradient is given by:
\begin{equation}
    \label{eq:expection_gradient}
    \nabla_{\theta}J(\theta) = \mathbb{E}_{x \sim \mathcal{D}, y \sim \pi_{\theta}(\cdot|x)}[(R(x, y) - b) \nabla_{\theta} \log \pi_{\theta}(y|x)].
\end{equation}
where $R(x, y)$ is the scalar reward assigned to the response $y$ for the prompt $x$. The term $A(x, y) = R(x, y) - b$ is known as the advantage function. The optimal baseline that minimizes variance is the true value function $V_{\pi}(x) = \mathbb{E}_{y \sim \pi_{\theta}(\cdot|x)}[R(x, y)]$, which represents the expected reward for a given state $x$. In practice, $V_{\pi}(x)$ is unknown and is often approximated by a learned critic network, denoted as $V_{\phi}(x)$.
However, a major challenge with this approach is the high variance of the gradient estimator, as the raw reward $R(x, y)$ can fluctuate significantly, leading to unstable training. To reduce this variance, a state-dependent baseline $b(x)$ can be subtracted from the reward without introducing bias into the gradient estimate. The resulting gradient estimator has provably lower variance:
To obtain a stable estimate of the advantage function, Generalized Advantage Estimation (GAE) is commonly employed. GAE calculates the advantage $\hat{A}_t$ using an exponentially weighted average of multi-step temporal difference (TD) errors, thereby effectively balancing the trade-off between bias and variance. The advantage is calculated by:
\begin{equation}
    \hat{A}_{t}^{\text{GAE}} = \sum_{l=0}^{\infty}(\gamma\lambda)^{l}\delta_{t+l},
\end{equation}
where $\delta_{t} = r_t + \gamma V_{\phi}(s_{t+1}) - V_{\phi}(s_t)$ and the parameter $\lambda \in [0, 1]$ controls the degree of bias and variance. Building on GAE, Proximal Policy Optimization (PPO) primarily aims to improve the sample efficiency of on-policy methods by allowing multiple gradient updates on a single batch of data~\cite{schulman2017proximal}. This off-policy capability is enabled by importance sampling, which uses a probability ratio $\rho_t(\theta) = \frac{\pi_{\theta}(a_t|s_t)}{\pi_{\theta_{old}}(a_t|s_t)}$ to correct for the distributional shift between the policy being updated and the one that generated the data. To prevent destructively large updates, PPO introduces its signature clipped scheme, as shown below:
\begin{equation}
    \hat{A}_t = \min(\rho_t(\theta)\hat{A}_{t}^{\text{GAE}}, \text{clip}(\rho_t(\theta), 1-\epsilon, 1+\epsilon)\hat{A}_{t}^{\text{GAE}}).
\end{equation}

While the clip scheme enhances stability, it introduces its own set of challenges.
The clipping threshold $\epsilon$ risks over-restricting policy exploration, potentially preventing the agent from discovering the optimal policy.
Moreover, some research suggests that even with this mechanism, PPO's performance can still be unpredictably unstable in certain tasks.
In light of these concerns and to prioritize robust and stable learning dynamics, we adopt a strict on-policy PPO for our experiments, where data is freshly collected with the latest policy before each optimization step.

\subsection{The Potential of Learning From Interaction}

The dominant paradigm for complex reasoning tasks samples $y \sim \pi_\theta(\cdot|x)$ comprising both an entire plan and final answer, where the entire trajectory is produced without external feedback. However, this approach compels the model to internally simulate a vast solution space, performing continuous enumeration, hypothesis generation, and validation. While this paradigm exhibits strong reasoning capabilities on tasks such as mathematics, code, and logic, it does so at the cost of low efficiency, often involving substantial ineffective exploration during the extended reasoning process. 

To address this limitation, we shift from generating an entire plan $y$ in one step to producing a sequence of thinking steps $T_t$ and action sets $A_t$ over multiple turns, where $A_t = \{a_t^1, a_t^2, \dots, a_t^{k_t}\}$ may contain multiple individual actions to lower cognitive burden. Each action set $A_t$ is followed by an environmental observation $o_t$. The maximum number of interaction turns is constrained by a predefined limit $L_{\text{max}}$. At each turn $t$, the model generates a thinking step $T_t$ and an action set $A_t \sim \pi_\theta(\cdot | x, H_{<t})$, where $H_{<t} = (o_1, T_1, A_1, o_2, \dots, o_{t-1}, T_{t-1}, A_{t-1})$ represents interaction history. The final response is the concatenation of all thinking steps, action sets, and observations, forming the complete trajectory $y = (o_1, T_1, A_1,o_2 \dots, o_L, T_L, A_L)$ and the optimization is to maximize the expected reward $J(\theta)$, implemented using the policy gradient framework defined in ~\cref{eq:expection_gradient}.
This interactive loop offers several key advantages.
First, environmental feedback provides an immediate corrective signal, enabling the model to rectify erroneous assumptions and thereby conserving computational resources that would otherwise be spent on flawed reasoning paths.
Second, learning from a series of feedback-driven interactions rather than a single lengthy and noisy trajectory improves training efficiency. Third, we argue that interaction facilitates superior internalization efficiency, allowing the model to more effectively simulate environmental dynamics and construct an internal world model.
\begin{figure}
    \centering
    \includegraphics[width=\linewidth]{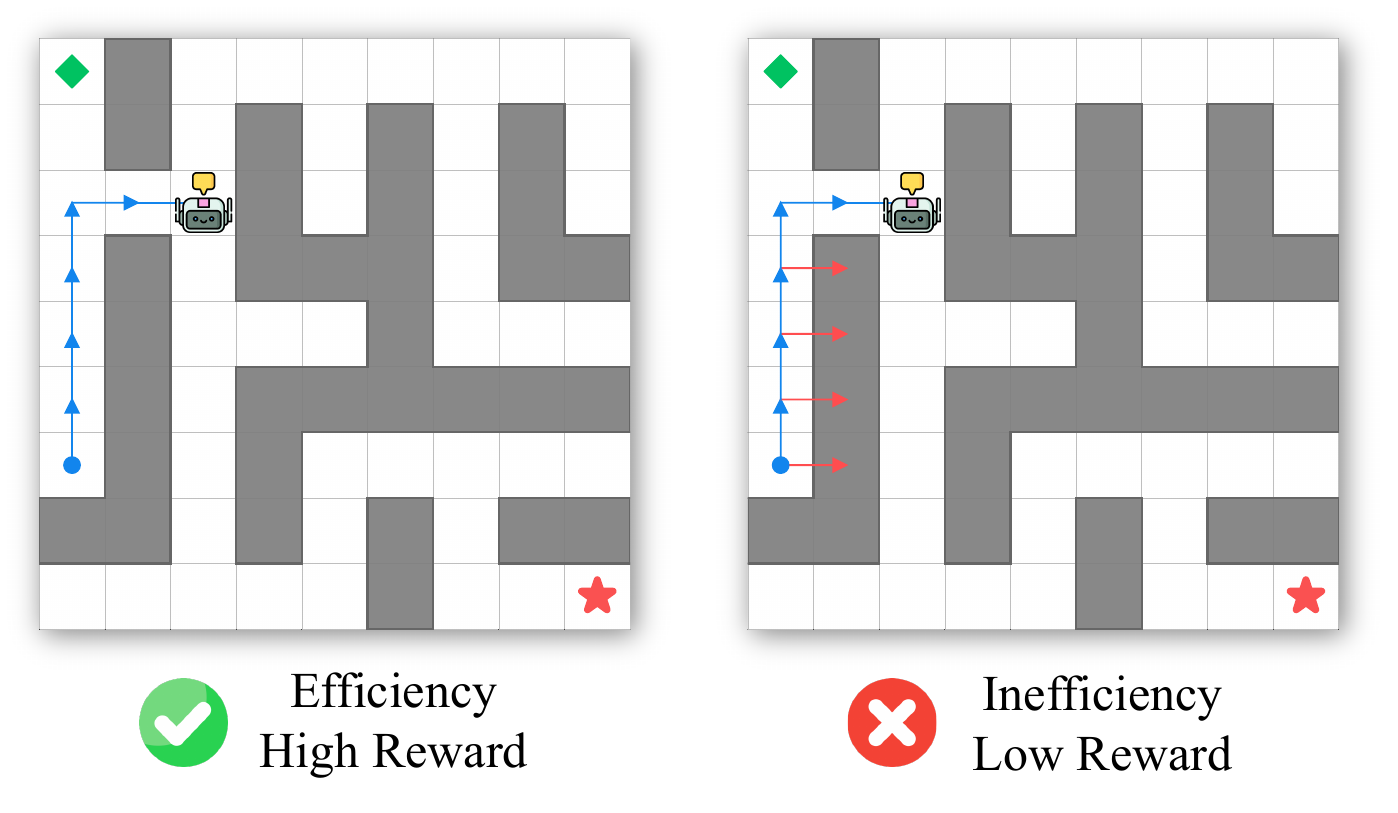}
    \caption{Comparison between different interactive strategies.}
    \label{fig:inefficent}
    \vspace{-1.3em}
\end{figure}
\subsection{\methodabb}
We formally present our \methodfull~(\methodabb), specifically designed to build robust world model reasoning via multi-turn interaction without constraining the model’s cognitive flexibility. Our approach addresses the challenges associated with several key aspects: (1) It introduces a reward scaling mechanism to ensure the model learns to effectively capitalize on immediate feedback. (2) It employs an interaction frequency annealing strategy to effectively manage the balance between reasoning and interaction, thus preventing the model from tending toward excessive interaction.

\noindent\textbf{Reward rescaling.}
Outcome-based rewards have proven effective in the math and code domain. However, in an interactive setting, relying solely on sparse outcome rewards can lead to excessively large exploration spaces. We observe that agents often resort to brute-force strategies, enumerating a large number of actions to complete a task shown in Figure~\ref{fig:inefficent}, rather than learning an efficient interactive strategy. This behavior results in low sample efficiency and poor generalization. To encourage more effective interaction strategies, we propose scaling the outcome reward by the proportion of effective actions within an episode. An action is considered effective if it leads to a state that is different from the previous one. Formally, if an episode contains $N$ actions and $N_{\text{eff}}$ of them are effective, the scaled reward $R_{\text{scaled}}$ is computed as:
\begin{equation}
\label{eq:scale_reward}
R_{\text{scaled}} = R_{\text{outcome}} \times \frac{N_{\text{eff}}}{N}.
\end{equation}
This scaling factor incentivizes the agent to minimize redundant or ineffective actions and to prioritize actions that yield meaningful progress. Since the effectiveness criterion is derived directly from state changes provided by the environment, it is inherently difficult for the agent to game the system or exploit the reward mechanism, ensuring that the reward signal remains grounded in genuine task progress.

\noindent\textbf{Interaction Frequency Annealing.}
While multi-turn interaction provides rich feedback, it also carries the risk of the model overly relying on environmental cues at the expense of internal reasoning, leading to shallow policies that fail to build a robust internal belief. To balance the immediate benefits of environmental feedback with the long-term need for deliberate world model reasoning, we introduce an interaction frequency annealing strategy. Specifically, for every $\tau$ training iteration, we compute two statistics over recent episodes: the average number of interaction turns per prompt, $\bar{L}$, and the maximum number of turns observed, $L_{\text{max}}'$. We then set the maximum allowed interaction turns for the subsequent training phase as follows:

\begin{equation}
    L_{\text{max}} = \frac{\bar{L} + L_{\text{max}}'}{2}.
\end{equation}

This dynamically adjusted limit serves two purposes. First, it adapts to the model’s capability, allowing fewer interactions as the policy becomes more efficient. Second, it implicitly implements a form of curriculum learning: early in training, the agent is permitted more turns to explore, while later stages encourage more efficient problem-solving within tighter constraints. The annealing process ensures the model gradually internalizes environmental dynamics.
\begin{table*}[!htbp]
\centering
\caption{Evaluation results on 3 tasks. 
\textbf{Standard}: Tasks with parameters identical to the training distribution. \textbf{Hard}: Tasks with the same environment type but more challenging environmental parameters. \textbf{PPO-EntirePlan} is a model trained to output the entire solution plan in a single turn which is common in RLVR of LLM. 
\textbf{PPO-Interactive} allows models to interact with the environment during training.
}
\label{tab:main_result}
\vspace{-0.5em}
\begin{tabular}{l l ccc cc cc}
\toprule
\multicolumn{2}{l}{\multirow{2}{*}{\textbf{Method}}} & \multicolumn{3}{c}{\textbf{Sokoban}} & \multicolumn{2}{c}{\textbf{Maze}} & \multicolumn{2}{c}{\textbf{Taxi}} \\
\cmidrule(lr){3-5} \cmidrule(lr){6-7} \cmidrule(lr){8-9}

\multicolumn{2}{c}{} &  \textbf{Standard} & \textbf{Hard-1} &  \textbf{Hard-2} & \textbf{Standard} & \textbf{Hard} & \multicolumn{2}{c}{\textbf{Standard}}  \\
\midrule
\textbf{\textit{Proprietary Models}} \\
\midrule
GPT-4o~\cite{hurst2024gpt} & &1.95 & 0 & 0 &4.69 & 1.56 &\multicolumn{2}{c}{6.25} \\
GPT-5~\cite{openaigpt5} & & 96.88&  91.41 & 87.89 & 99.61 & 97.66 & \multicolumn{2}{c}{36.72} \\
OpenAI o3~\cite{openaio3} & & 98.83& 90.23 & 85.16 & 95.7& 93.36& \multicolumn{2}{c}{72.66} \\
OpenAI o4-mini~\cite{openaio3} &  &83.20& 69.14 &  62.89&93.75& 78.52& \multicolumn{2}{c}{83.98} \\
Gemini 2.5 Pro~\cite{comanici2025gemini} & & 36.72 & 17.19 & 20.31& 83.59 & 63.28 & \multicolumn{2}{c}{17.97} \\
Claude 4.5 Sonnet~\cite{claude4p5}& & 60.94& 34.77& 29.3 & 94.14 & 68.36& \multicolumn{2}{c}{66.41} \\
\midrule
\textbf{\textit{Opensource Models}} \\
\midrule
Qwen2.5-7B-Instruct~\cite{yang2024qwen2.5}& & 0 & 0 & 0 &  0& 0 & \multicolumn{2}{c}{1.17} \\
Qwen2.5-32B-Instruct~\cite{yang2024qwen2.5}& & 0& 0& 0 & 0.68 & 0.39 & \multicolumn{2}{c}{2.34} \\
Qwen3-8B~\cite{qwen3technicalreport}& & 16.41& 4.60& 1.17 & 63.67 & 17.76 & \multicolumn{2}{c}{9.38} \\
Qwen3-14B~\cite{qwen3technicalreport}& & 18.75& 6.64& 5.08 & 72.27 & 28.52 & \multicolumn{2}{c}{6.64} \\
\midrule
\textbf{\textit{Method Comparison}} \\
\midrule
Qwen3-8B-Own& & 3.29 & 0.84 & 1.39 &1.95 &0.20 & \multicolumn{2}{c}{5.60} \\
PPO - EntirePlan  & & 49.12 & 2.34& 0.35 &  75.04 & 26.51& \multicolumn{2}{c}{38.92} \\
PPO - Interactive  & &64.21 &41.26 &46.83  &83.74  & 36.52 & \multicolumn{2}{c}{39.16} \\
\rowcolor{mygray}
\methodabb   &  & 78.57 & 52.68 & 49.90 &88.14 & 50.59 & \multicolumn{2}{c}{62.16} \\
\bottomrule
\end{tabular}
\end{table*}

\section{Experiments}
\label{sec:exp}
\begin{figure*}
    \centering
    \includegraphics[width=\linewidth]{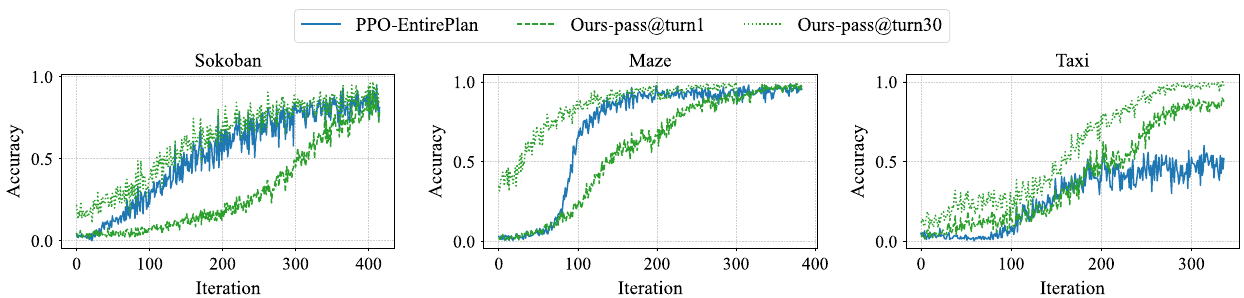}
    \caption{Training accuracy curves comparing PPO-EntirePlan, \methodabb's single-turn and multi-turn performance. The single-turn performance of \methodabb~ not only is comparable to the PPO-EntirePlan but also gradually approaches its own multi-turn training performance across all environments. The Taxi environment \methodabb~exemplifies a clear breakthrough beyond the PPO-EntirePlan performance ceiling.}
    \label{fig:env_curve}
    \vspace{-1.0em}
\end{figure*}
\subsection{Settings}
We validate our \methodabb~on three distinct environments: Taxi, Maze, and Sokoban. The Taxi environment involves navigation and passenger pickup/dropoff tasks, designed to test the agent's ability to follow complex instructions and handle sequential dependencies. The Maze environment requires spatial navigation through a grid world, evaluating spatial reasoning and pathfinding capabilities. The Sokoban environment presents a puzzle-solving task that demands irreversible symbolic planning and foresight, challenging the model's ability to reason about long-term action consequences. More details are provided in the Appendix.

\noindent\textbf{Training settings.}
We employ a strict on-policy PPO algorithm for training, which omits the KL divergence loss and the entropy bonus.
We primarily conduct our experiments using Qwen3-8B-Own, a model subject to supervised fine-tuning on our proprietary general and reasoning datasets, with Qwen2.5-instruct-7B~\cite{yang2024qwen2.5} also included for comparison to isolate the effect of base model capabilities.
During training, the global batch size is set to $256$. We sample 8 responses per prompt using a decoding temperature of $1.0$. The maximum completion length is set to $16$k tokens, with an initial maximum limit of $30$ interaction turns.

\noindent\textbf{Evaluation settings.}
We assess the performance of our method under two distinct evaluation protocols: (1) Standard: in-domain evaluation on tasks of the same difficulty level as those encountered during training; (2) Hard: in-domain evaluation on tasks with increased difficulty to test for robustness and generalization; For Sokoban, there are two adjustable environmental parameters: the number of boxes and the map size. Hard-1 was constructed by increasing the original map size by half while maintaining the same number of boxes, and Hard-2 was created by increasing the number of boxes while keeping the map size unchanged. For Maze, we constructed Hard-1 by increasing the original map size by half. We repeat eight evaluations on a fixed set of $256$ held-out problems generated from the corresponding environment and calculate the average as a result to mitigate randomness, and all results are evaluated in a single-turn manner for fair comparison. 
Additionally, we conduct a comprehensive evaluation of our paradigm on several representative benchmarks spanning three categories: General (MMLU-Pro~\cite{wang2024mmlupro}, GPQA-Diamond~\cite{rein2024gpqa}, LiveBench~\cite{white2024livebench}), Math (AIME24~\cite{AoPS_AIME}, AIME25~\cite{AoPS_AIME}, BeyondAIME~\cite{bytedance_seed_2025_beyondaime}, HMMT25~\cite{HMMT2025}), and Code (LiveCodeBench v5~\cite{jain2025livecodebench}).
We conducted multiple $\text{pass}@1$ estimations to ensure evaluation stability.

\begin{table*}[htbp]
\centering
\caption{Performance across key benchmarks. Results marked with $\dagger$ are directly cited from the original paper or technical report.}
\label{tab:general_bench}
\vspace{-0.6em}
\resizebox{\textwidth}{!}{
\begin{tabular}{lcccccccc}
\toprule
\multirow{2}{*}{\textbf{Model}} & \multicolumn{4}{c}{\textbf{AVG@64}} & \multicolumn{2}{c}{\textbf{AVG@16}} & \multicolumn{1}{c}{\textbf{AVG@4}} & \multicolumn{1}{c}{\textbf{AVG@1}} \\
\cmidrule(lr){2-5} \cmidrule(lr){6-7} \cmidrule(lr){8-8} \cmidrule(lr){9-9}
 & \textbf{AIME24}~\cite{AoPS_AIME} & \textbf{AIME25}~\cite{AoPS_AIME} & \textbf{BeyondAIME}~\cite{bytedance_seed_2025_beyondaime} & \textbf{HMMT25}~\cite{HMMT2025} & \textbf{GPQA-Diamond}~\cite{rein2024gpqa} & \textbf{LiveCodeBench v5}~\cite{jain2025livecodebench}  & \textbf{LiveBench}~\cite{white2024livebench} & \textbf{MMLU-PRO}~\cite{wang2024mmlupro} \\
\midrule
OpenAI o1$^\dagger$~\cite{openaio1} & 74.30 & 79.20 & - & - & 78.00 & 63.90 & 75.70 & 89.30 \\
OpenAI o3-mini$^\dagger$~\cite{openaio3} & 79.60 & 74.80 & 63.60 & - & 76.80 & 66.30 & 70.00 & 78.70 \\
GPT-4o$^\dagger$~\cite{hurst2024gpt} & 11.10 & 7.60 & - & - & 46.00 & 32.70 & 52.20 & 72.55 \\
Qwen3-14B Thinking$^\dagger$~\cite{qwen3technicalreport} & 79.30 & 70.40 & - & - & 64.00 & 63.50 & 71.30 & - \\
\midrule
Qwen3-8B-Own & 85.10 & 77.92 & 52.88 & 63.44 & 59.91 & 65.32 & 67.93 & 72.35 \\
\rowcolor{mygray}
WMAct-Sokoban & 86.56 & 79.48 & 55.14 & 68.49 & 62.15 & 67.14& 69.60 & 73.14 \\
\bottomrule
\end{tabular}
}
\vspace{-1em}
\end{table*}

\subsection{Main Results}
\noindent\textbf{Performance on standard tasks}. We first evaluate the performance of \methodabb~ on standard tasks with difficulty levels identical to those in the training distribution. As shown in Table~\ref{tab:main_result}, \methodabb~ achieves superior results compared to all baseline methods across the Sokoban, Maze, and Taxi environments. Specifically, \methodabb~ attains success rates of 78.57 on Sokoban, 88.14 on Maze, and 62.16 on Taxi. These scores substantially outperform both the PPO-EntirePlan and the standard multi-turn PPO approach. Notably, \methodabb~ even surpasses the performance of several much larger proprietary models such as GPT-4o and Claude 4.5 Sonnet on these standard tasks, demonstrating the effectiveness of our interactive learning paradigm coupled with efficacy-driven exploration and frequency annealing.

\noindent\textbf{Generalization on challenging tasks}.
To assess the robustness and generalization capability of our method, we further evaluate \methodabb~on harder task variants that feature increased complexity. As presented in Table~\ref{tab:main_result}, \methodabb~maintains strong performance in these challenging settings, achieving 52.68 on Sokoban Hard-1, 49.90 on Sokoban Hard-2, and 50.59 on Maze Hard. This represents a significant improvement over the PPO-EntirePlan and PPO-Interactive, which exhibit a considerable performance drop on these harder tasks.
The results indicate that \methodabb~ effectively internalizes environmental dynamics through multi-turn interaction, enabling it to generalize to unseen and more complex scenarios without overfitting to the training distribution.
The ability to condense learned knowledge into a robust internal world model allows \methodabb~to solve tasks efficiently even under tightened constraints and increased difficulty.

\noindent\textbf{Efficient world model reasoning}.
Figure~\ref{fig:env_curve} presents the training accuracy curves for the Sokoban, Maze, and Taxi environments under three experimental settings: the PPO-EntirePlan, \methodabb's single-turn and multi-turn accuracy. 
Critically, the single-turn performance of our method is not only comparable to the PPO-EntirePlan but also gradually approaches its own multi-turn training performance across all environments. This convergence provides strong evidence that the knowledge acquired through interaction is effectively compressed and internalized, enabling robust performance even without environmental feedback. 
Furthermore, in the Taxi environment, both the multi-turn and single-turn accuracies of our method clearly break through the performance ceiling established by the PPO-EntirePlan baseline. This result underscores that our interactive learning paradigm not only facilitates more efficient exploration during training but also yields a final policy that is more competent and generalizable, solidifying the overall effectiveness of our approach.

\noindent\textbf{Performance on General Benchmarks.} To confirm that the skills learned during interactive training are transferable, we evaluated \methodabb~model trained with the Sokoban environment on a broad range of general benchmarks. As shown in Table~\ref{tab:general_bench}, our WMAct-Sokoban model demonstrates comprehensive improvements over the Qwen3-8B-Own model, which shows its largest gains in complex reasoning, with a major 5.05 point increase on HMMT25 and a 2.24 point increase on GPQA-Diamond. The model also improved across general-purpose tasks, illustrated by a +1.67 point gain in LiveBench. This diverse range of improvements, from substantial leaps in math to steady gains in general tasks, provides strong evidence that our thinking by doing paradigm enhanced the reasoning and planning capability, allowing these skills to generalize far beyond the original interactive environment.

\subsection{Ablation Study}

\begin{table}[!t]
    \centering
    \caption{Ablation study of key components on Sokoban environment. The table demonstrates that reward rescaling and frequency annealing provide substantial and complementary improvements over the PPO-Interactive baseline, especially in harder settings.}
    \label{tab:ablation_component}
    \vspace{-0.6em}
    \resizebox{\columnwidth}{!}{
    \begin{tabular}{l ccc} 
    \toprule
    \multirow{2}{*}{\textbf{Method}} & \multicolumn{3}{c}{\textbf{Sokoban}} \\
    \cmidrule(lr){2-4} %
    & \textbf{Standard} & \textbf{Hard-1}& \textbf{Hard-2} \\
    \midrule
    \text {PPO-EntirePlan } & 49.12 & 2.34 & 0.35 \\
    \midrule
    \text {PPO-Interactive} & 64.21 & 46.83 & 41.26 \\
    \hspace{0.4em}\text { + reward rescaling } & 73.68 & 50.78 & 48.05\\
    \hspace{1em}\text { + frequency annealing } & \textbf{78.57} & \textbf{52.68} & \textbf{49.90}\\
    \bottomrule
    \end{tabular}}
    \vspace{-0.8em}
\end{table}

\noindent\textbf{The effects of each component.} To verify the effectiveness of the proposed method, we examine different combinations of components. In Table \ref{tab:ablation_component}, the introduction of interactive training PPO-Interactive substantially improves performance over the non-interactive baseline, particularly on harder task variants. This demonstrates the fundamental benefit of multi-turn interaction. Adding the proposed effective action reward scaling further boosts results significantly, with the most notable gains observed on the Hard-2 task, where performance improves from 41.26 to 48.05. This indicates the reward mechanism successfully incentivizes more efficient exploration by penalizing redundant actions. Incorporating interaction frequency annealing yields the best overall performance, pushing scores to 78.57, 52.68, and 49.90 across the three difficulty levels. The consistent improvement across all settings confirms the annealing strategy effectively balances immediate environmental feedback with the development of internal reasoning capabilities, enabling the model to gradually internalize environmental dynamics and maintain exploration efficiency.

\begin{table}[!t]
\centering
\caption{Performance comparison of \textbf{step penalty} and \textbf{frequency annealing} regularization methods on Sokoban environments, including an investigation into the effect of the annealing interval $\tau$.}
\label{tab:step_penalty_optimized_v2}
\vspace{-0.6em}
\resizebox{\columnwidth}{!}
{
\begin{tabular}{llccc} %
\toprule
\multirow{2}{*}{\textbf{Method}} & \multirow{2}{*}{\textbf{Param}} & \multicolumn{3}{c}{\textbf{Sokoban}} \\
\cmidrule(lr){3-5}
& & \textbf{Standard} & \textbf{Hard-1} & \textbf{Hard-2} \\
\midrule
{step penalty} & $ - 0.1$ & 72.43 & 49.32 & 45.46 \\ 
\midrule
\multirow{3}{*}{{frequency annealing}} & $\tau=50$ & 77.73 & 52.93 & 47.56 \\
& $\tau=100$ & 78.57 & 52.68 & 49.90 \\
& $\tau=150$ & 74.71 & 49.95 & 44.24 \\
\bottomrule
\end{tabular}
}
\vspace{-1.5em}
\end{table}

\noindent\textbf{Comparison between step penalty and frequency anneal}.
To mitigate the issue of inefficient trajectories, much of the early RL literature incorporates a small, fixed step penalty into the environmental reward. Table~\ref{tab:step_penalty_optimized_v2} compares the performance of two distinct regularizations, step penalty and interaction frequency annealing,  within the Sokoban environment. The frequency annealing achieves a success rate of 78.57 on the Standard level, surpassing the step penalty by 6.14, and maintains generalization on the more challenging levels. The step penalty, which is designed to incentivize path efficiency by imposing a constant negative reward, steers the agent toward myopic strategies, hindering global performance. Conversely, the frequency annealing, which modulates the agent's interaction frequency with the environment, provides a more effective route by facilitating a more thorough exploration early in training and promotes internalization of environmental dynamics, developing efficient world model reasoning.

\noindent\textbf{The effects of $\tau$.} We investigate the impact of the interaction frequency annealing interval $\tau$ on model performance. As shown in Table~\ref{tab:step_penalty_optimized_v2}, the optimal performance across all Sokoban task variants is achieved with a medium annealing interval of $\tau=100$. This setting yields the best scores of 78.57, 52.68, and 49.90 for Standard, Hard-1, and Hard-2 tasks, respectively. The performance degradation with shorter $\tau=50$ is particularly pronounced on Hard-2, which introduces additional boxes requiring more complex planning. This suggests that overly frequent annealing adjustments may disrupt the policy's ability to develop sophisticated multi-step reasoning. Conversely, the performance drop with longer $\tau=150$ affects all tasks but is most evident in Standard and Hard-1 environments, where the model fails to efficiently reduce its interaction dependency. This pattern reveals that the annealing schedule must balance two competing needs: providing sufficient interaction turns for complex skill acquisition while progressively encouraging internalization of environmental dynamics. The medium $\tau=100$ strikes this balance effectively, allowing adequate exploration time for Hard-2's increased complexity while maintaining efficiency gains in simpler environments.

\noindent\textbf{The effects of model prior.} 
\begin{figure}
    \centering
    \includegraphics[width=\linewidth]{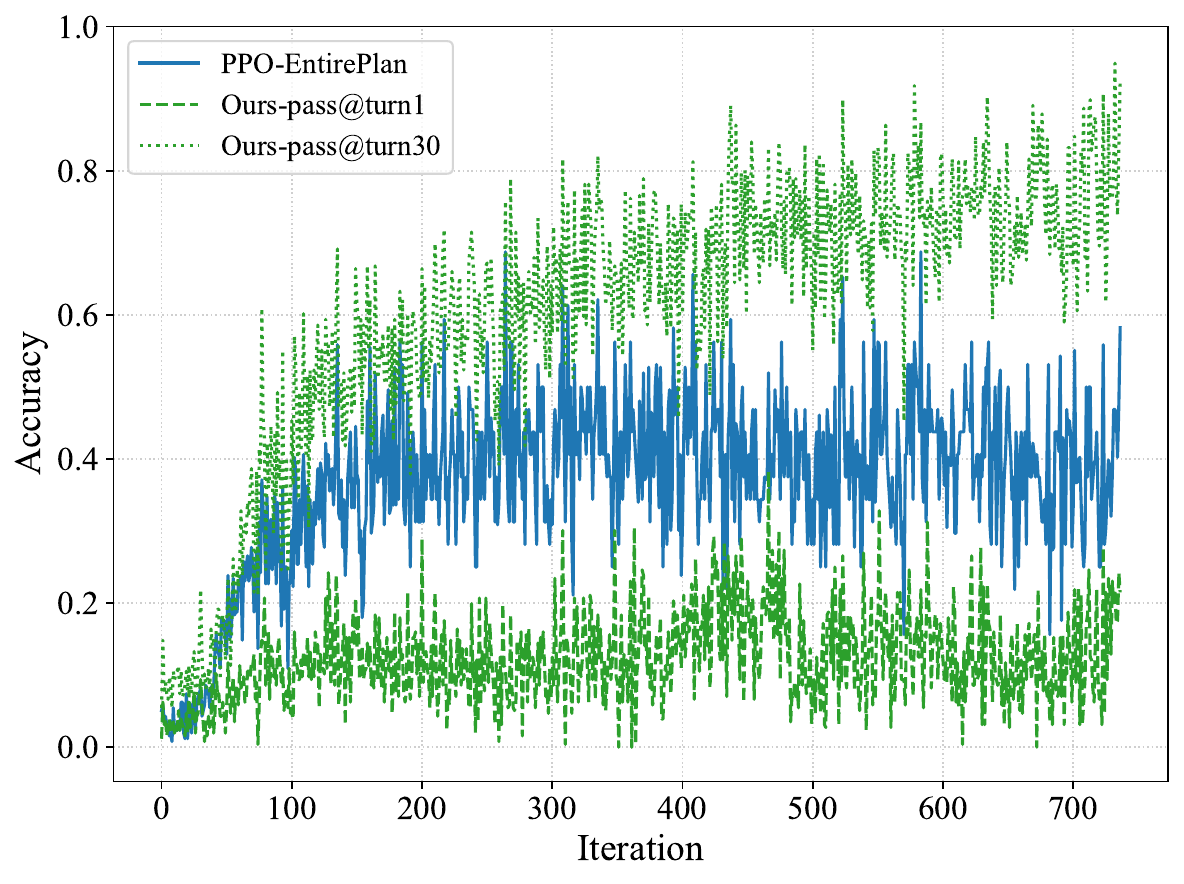}
    \vspace{-2em}
    \caption{The training dynamic of Qwen2.5-7B-Instruct. Qwen2.5-7B-instruct failed to internalize knowledge from multi-turn interactions to improve its problem-solving efficiency, which underscored the importance of advanced reasoning patterns.}
    \label{fig:qwen25}
    \vspace{-1.5em}
\end{figure}
We present training dynamics of Qwen2.5-instruct-7B in Figure~\ref{fig:qwen25}. A key observation is that the Qwen2.5-instruct-7B model does not exhibit the same trend of multi-turn interaction driving improved single-turn performance as seen in more advanced models. This suggests that Qwen2.5-instruct-7B fails to internalize environmental changes into its own belief state effectively. We attribute this primarily to the model's lack of advanced reasoning patterns such as reflection, self-correction, and strategic foresight. These patterns are crucial for internalization because they enable the model to abstract general principles from specific feedback, update its internal state meaningfully, and form compressed representations of environmental dynamics. Without these capabilities, the model treats each interaction as an isolated event rather than as evidence for refining a coherent world model. This highlights that architectural and behavioral patterns are as important as the interactive learning framework itself for achieving emergent world modeling capabilities. More visualization and statistics are presented in the Appendix.

\section{Discussion and Conclusion}
\label{sec:conclusion}
Our study illustrates \textit{how RL fundamentally shapes the development of world model reasoning}, moving models beyond rigid, human-designed cognitive patterns. We identify multi-turn interaction paradigm as a potent mechanism, while monolithic reasoning imposes a substantial cognitive burden.
The thinking-by-doing paradigm bypasses this burden by allowing agents to validate solution paths through direct environmental feedback, which effectively prunes incorrect strategies and reinforces environmental dynamics into reasoning.
However, we identify several critical constraints on this process:

\noindent\textbf{The Primacy of Cognitive Behaviors.} Advanced reasoning patterns, such as reflection or self-correction, potentially transferred from domains like mathematics or coding, are crucial for models to process feedback effectively, enabling them to link environmental outcomes to preceding actions. Without these capabilities, the model views interactions merely as isolated events rather than evidence to refine a coherent world model, ultimately hindering environmental dynamics modeling and world model reasoning.

\noindent\textbf{Interaction Strategy on Knowledge Acquisition.} We observed models adopting inefficient, brute-force strategies, attempting to enumerate a vast action space rather than analyzing optimal paths. This approach impairs the acquisition of high-quality knowledge from feedback. We address this limitation by implementing reward rescaling to incentivize efficacious exploration, thereby steering the model from simple enumeration toward purposeful interaction.

\noindent\textbf{The Risk of Feedback Over-Reliance.} The informational richness derived from environmental feedback may induce an over-reliance on interaction, which can stifle the development of robust, long-horizon world model reasoning, as models adopt myopic, reactive policies instead of internalizing environmental dynamics. To mitigate this, we employ an interaction frequency annealing strategy to compel the model to foster efficient long-horizon planning capabilities.

\clearpage
\appendix

\setcounter{page}{1}
\renewcommand{\thepage}{A-\arabic{page}}

\renewcommand{\thefigure}{\arabic{figure}}
\setcounter{figure}{0}
\renewcommand{\thetable}{\arabic{table}}
\setcounter{table}{0}

\section*{Supplementary Material}

\section{Experiment Setting}
\subsection{Environments}
We evaluate our method on three distinct, challenging grid-world environments—Maze, Sokoban, and Taxi—all requiring agents to execute long-horizon sequential planning and meticulous state tracking. These environments are specifically adapted to interface with Language Models (LLMs) by converting the internal state into a standardized text-based ASCII map.
The system also employs a textual action space with action parsing and validation, ensuring a predictable protocol for all environments.
The complete prompt of the first turn provided to the LLM at each step follows a standardized structure: \texttt{\{system prompt\} \{environment description\} \{environment action prompt\}}, where specific examples are shown in Section~\ref{sec:prompt}.
Subsequent turns only contain the description of the environment feedback and the action prompt.

\begin{figure}[htb!] %
    \centering

    \begin{subfigure}[t]{0.45\textwidth}
        \centering
        \includegraphics[width=\linewidth]{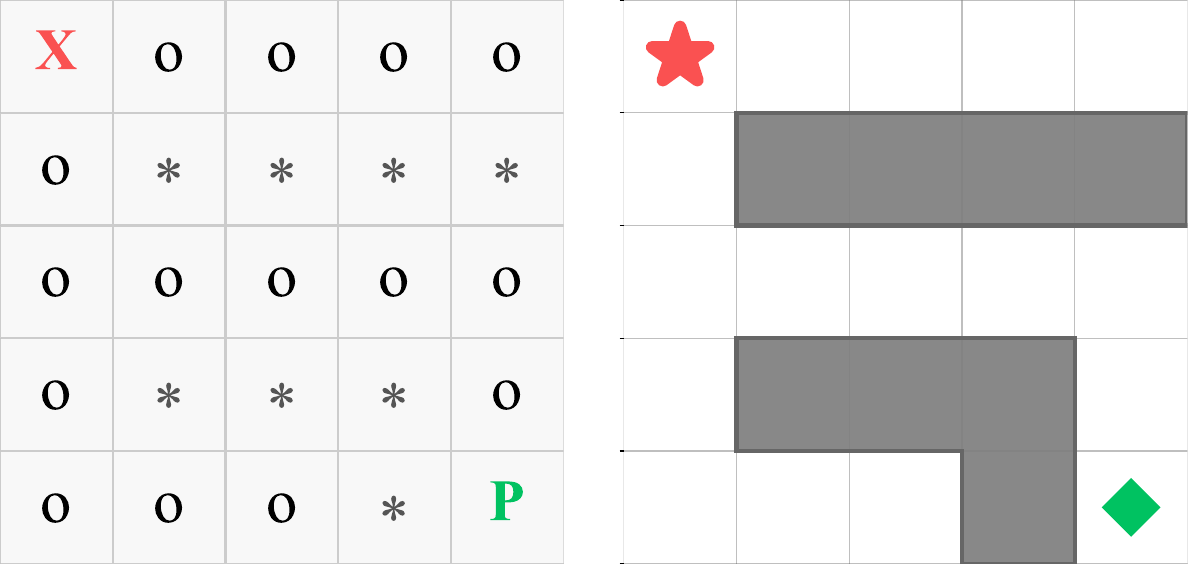} 
        \caption{Maze Environment: Topological Navigation \& Pathfinding}
        \label{fig:sub_maze}
    \end{subfigure}%
    \vspace{1em}
    
    \begin{subfigure}[t]{0.45\textwidth}
        \centering
        \includegraphics[width=\linewidth]{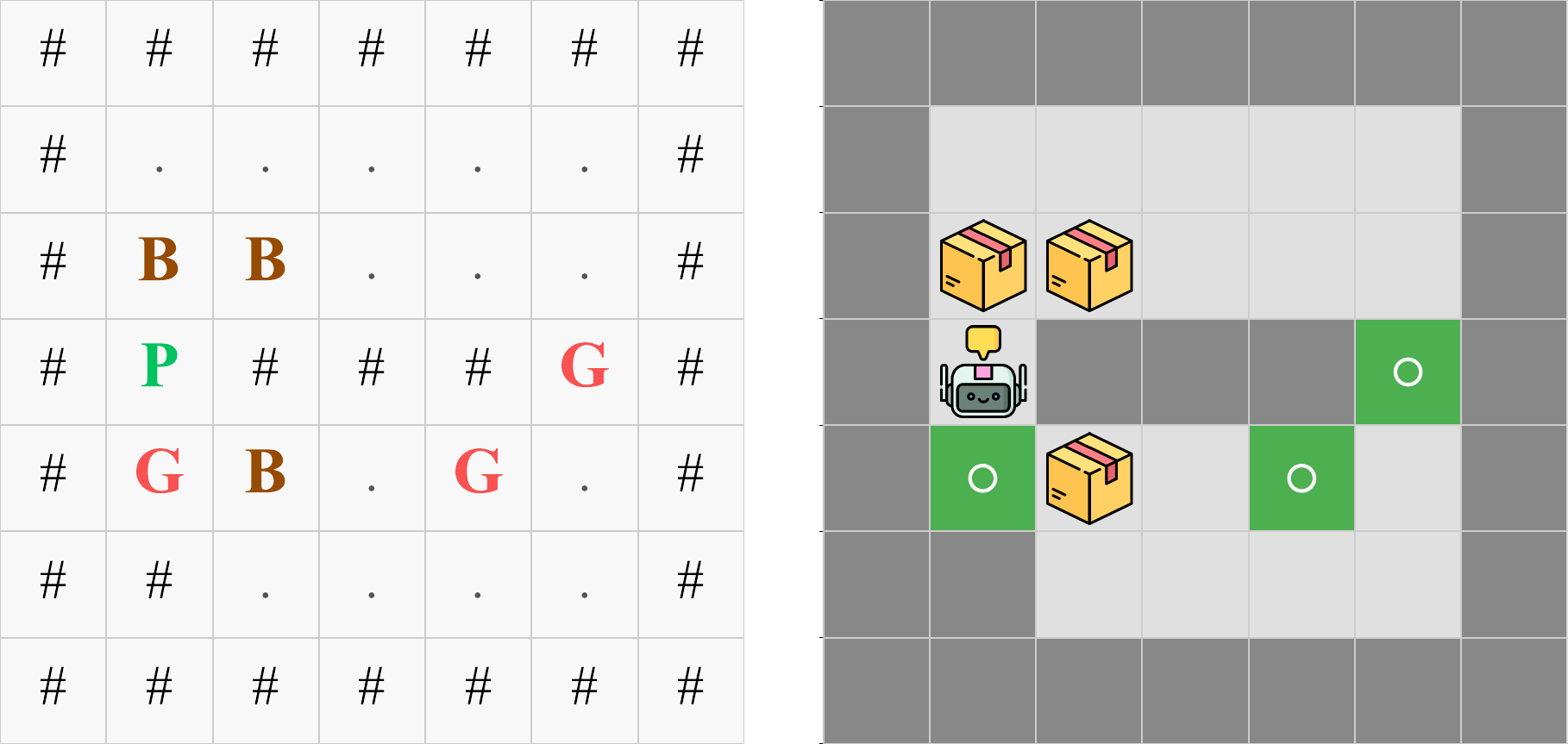} 
        \caption{Sokoban Environment: Object Manipulation \& Deadlocks}
        \label{fig:sub_sokoban}
    \end{subfigure}%
    \vspace{1em}
    
    \begin{subfigure}[t]{0.45\textwidth}
        \centering
        \includegraphics[width=\linewidth]{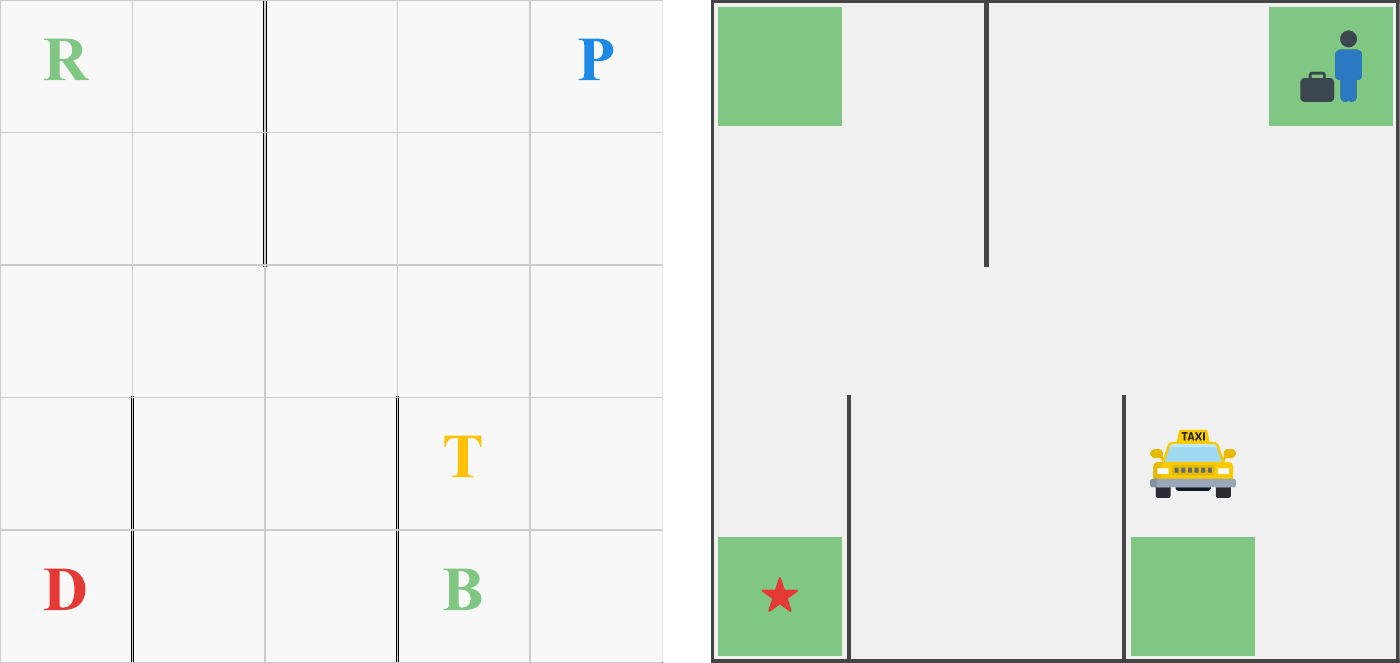} 
        \caption{Taxi Environment: Sequential Decision-making}
        \label{fig:sub_taxi}
    \end{subfigure}

    \caption{\textbf{Environment Visualization.}
    For each environment, the left panel shows the standardized character map for agent observation, and the right panel presents the visual illustration for human intuition and inspection.}
    \label{fig:3_subfigs}
    \vspace{-1.5em}
\end{figure} %

The Maze task (Figure~\ref{fig:sub_maze}) focuses on topological navigation and pathfinding under structural constraints where complexity is intrinsically linked to the grid size. Our configuration specifically utilizes $11 \times 11$ and $15 \times 15$ boards to span Standard and Hard difficulty regimes across all stages of experimentation. Maze construction employs the PRIM algorithm, which is favored for its generation of structurally intricate and highly varied topological layouts, an advantage over the DFS method's tendency to yield repetitive, long-corridor environments. Upon initialization, the agent begins at an available corner cell, and the destination is programmatically designated as the path point maximally distant from the starting location, thus guaranteeing a maximal planning distance for each instance. The environment state is then rendered to the LLM as the ASCII map employing dedicated symbols: P for the player, X for the goal, o for navigable paths, and * for walls. Interaction is mediated by five discrete actions—move up, down, left, right, and no operation—with movement strictly validated against structural constraints. The episode concludes either upon goal achievement or when the operational limit is reached.

The Sokoban environment (Figure~\ref{fig:sub_sokoban}) is the primary testbed for object manipulation and deadlocks due to its inherent requirement for non-reversible action constraints, requiring the agent to push all boxes onto designated goal locations. Task complexity is precisely controlled by two orthogonal factors: the grid dimensionality and the number of boxes. Our configuration defines three distinct difficulty regimes: Standard ($7 \times 7$ grid, $2$ boxes), Hard-1 (${10 \times 10}$ grid, ${2}$ boxes, increasing the spatial search space), and Hard-2 (${7 \times 7}$ grid, ${3}$ boxes, increasing object coordination complexity). All generated puzzle instances are guaranteed to be solvable, achieved by utilizing a mechanism that generates the starting state through a controlled backtracking sequence from a solved state. The environment state is presented to the LLM as the ASCII map, leveraging clear symbols for all entities: P for the player, B for the box, G for the goal, * for the box on goal, \# for walls, and . for empty spaces. Interaction is mediated by a comprehensive discrete action space that critically separates simple positional movement—move up, down, left, right—from the core object manipulation commands—push up, down, left, right—alongside a necessary no operation command.

\begin{table*}[!ht]
\centering
\caption{More results across general benchmarks.}
\label{tab:sup_general_bench}
\resizebox{\textwidth}{!}{
\begin{tabular}{lcccccccc}
\toprule
\multirow{2}{*}{\textbf{Model}} & \multicolumn{4}{c}{\textbf{AVG@64}} & \multicolumn{2}{c}{\textbf{AVG@16}} & \multicolumn{1}{c}{\textbf{AVG@4}} & \multicolumn{1}{c}{\textbf{AVG@1}} \\
\cmidrule(lr){2-5} \cmidrule(lr){6-7} \cmidrule(lr){8-8} \cmidrule(lr){9-9}
 & \textbf{AIME24}~\cite{AoPS_AIME} & \textbf{AIME25}~\cite{AoPS_AIME} & \textbf{BeyondAIME}~\cite{bytedance_seed_2025_beyondaime} & \textbf{HMMT25}~\cite{HMMT2025} & \textbf{GPQA-Diamond}~\cite{rein2024gpqa} & \textbf{LiveCodeBench v5}~\cite{jain2025livecodebench}  & \textbf{LiveBench}~\cite{white2024livebench} & \textbf{MMLU-PRO}~\cite{wang2024mmlupro} \\
\midrule
Qwen3-8B-Own & 85.10 & 77.92 & 52.88 & 63.44 & 59.91 & 65.32 & 67.93 & 72.35 \\
WMAct-Maze & 85.21 & 78.07 & 53.75 & 65.10 & 61.68 & 67.56 & 68.54 & 72.64 \\
WMAct-Taxi & 85.00 & 78.80 & 53.02 & 65.57 & 61.24 & \textbf{67.70} & 69.14 & 72.71\\
WMAct-Sokoban & \textbf{86.56} & \textbf{79.48} & \textbf{55.14} & \textbf{68.49} & \textbf{62.15} & 67.14& \textbf{69.60} & \textbf{73.14} \\
\bottomrule
\end{tabular}
}
\end{table*}

The Taxi environment (Figure~\ref{fig:sub_taxi}) is a classic reinforcement learning domain designed to assess an agent's ability to perform sequential decision-making, meticulous state tracking, and multi-step sub-goal planning—pickup, navigation, and dropoff—within a discrete $5 \times 5$ grid world. The taxi agent is tasked with retrieving a passenger from one of four designated initial locations and delivering them to one of four specified destinations. 
The complexity is determined by the state space size and the strictness of the sequential rules, involving a fixed execution order: picking up the passenger first and then dropping them off at the correct destination.
The environment state is presented to the LLM as the ASCII map, which visually renders the geometric layout and utilizes a detailed legend for all entities: T for the taxi, P for the passenger, D for the destination, and critical compounded symbols like T(P) for the taxi carrying the passenger and T/P for the taxi and unpicked passenger co-located. 
The action space comprises six fundamental commands: four directional navigation actions—move up, down, left, and right—and two interaction actions, including pickup and dropoff. These interaction actions are subject to strict validity checks (e.g., dropoff is only valid at the destination while carrying the passenger), with invalid attempts resulting in immediate negative feedback.

\subsection{Training Setting}
We conduct reinforcement learning on instances of our Qwen-8B-Own and Qwen-2.5-Instruct. We sample 256 trajectories from 32 distinct environment prompts, each replicated 8 times, to increase prompt efficiency.
Specifically, we use a vLLM backend configured with a temperature of 1.0, top-$p$ of 1.0, and top-$k$ of 0. Maximum generation length is constrained within 16K tokens, with a maximum of 12K tokens per turn. Each turn, the generated response will contain a think section and several actions, wrapped in \texttt{\textless{}think\textgreater{}\textless{}/think\textgreater{} } and \texttt{\textless{}action\textgreater{}\textless{}/action\textgreater{}} tags, respectively. We will extract and sequentially execute all actions found within the \texttt{\textless{}action\textgreater{}\textless{}/action\textgreater{}} tags.
The overall reward $R_{\text{overall}}$ is determined by an outcome reward and a format reward, where an invalid action results in a small penalty, ensuring operational integrity. For WMAct, the outcome reward is proportionally scaled based on Eq~\ref{eq:scale_reward}.
And then, we employ the on-policy Proximal Policy Optimization (PPO) algorithm on entire trajectories, where generalized Advantage Estimation (GAE) with $\gamma=1$ and $\lambda=1$ is used to evaluate the advantage. The KL penalty and entropy regularization are disabled. The policy and critic networks are optimized using Adam with learning rates of $1\times10^{-6}$ and $5\times10^{-6}$, respectively. These learning rates are linearly warmed up over the first 20 and 50 iterations. We perform one policy update for every 12 critic updates, with no initial critic warm-up phase.

\subsection{Evaluation Setting}
We held out 256 prompts as an evaluation set for each environment to ensure they would not appear in the training set.
Our evaluation protocol mirrors the training configuration to ensure consistent assessment where the vLLM-based decoding stack (temperature=1.0, top-$p$=1.0, top-$k$=0) and enforces the same think-action response structure.
An evaluation episode concludes upon the first of these conditions being met: (1) the environment signals task completion, (2) the generation length reaches a predefined maximum, or (3) the context token budget is nearly exhausted. 
\section{More Evaluation Results}

\subsection{Supplemental General Benchmarks}
As a complementary analysis, Table $\ref{tab:sup_general_bench}$ further compares the general capabilities of models trained across three environments of distinct complexity: Maze, Sokoban, and Taxi.
The results clearly indicate that while all models trained with the interactive WMAct paradigm exhibit significant improvement over the base model Qwen3-8B-Own, the inherent difficulty and challenge type of the training environment critically influence the generalization of the learned skills. Specifically, the WMAct-Sokoban maintains its lead across most metrics, demonstrating particular strength in complex reasoning and state tracking benchmarks such as HMMT25 (68.49) and GPQA-Diamond (62.15). This strongly supports our view that the requirements of Sokoban—specifically dealing with non-reversible actions and deadlock avoidance—compel the model to learn a more refined, abstract planning capability with long-term dependency tracking. Conversely, the performance of WMAct-Maze and WMAct-Taxi is generally strong but slightly inferior to the Sokoban-trained agent. As the Maze primarily focuses on topological pathfinding and the Taxi emphasizes multi-step sub-goal decomposition, the reasoning constraints in these two tasks are relatively less stringent. However, we observe a notable exception on the LiveCodeBench v5 metric, where WMAct-Taxi achieves a slight lead of 67.70 over WMAct-Sokoban (67.14). This marginal advantage may be attributable to the Taxi environment's core requirement of sequential sub-goal planning, a structure that closely mirrors the execution flow and step-by-step function calling often required in code generation and problem-solving benchmarks like LiveCodeBench v5.

\begin{figure*}[h!]
    \centering
    \includegraphics[width=1\linewidth]{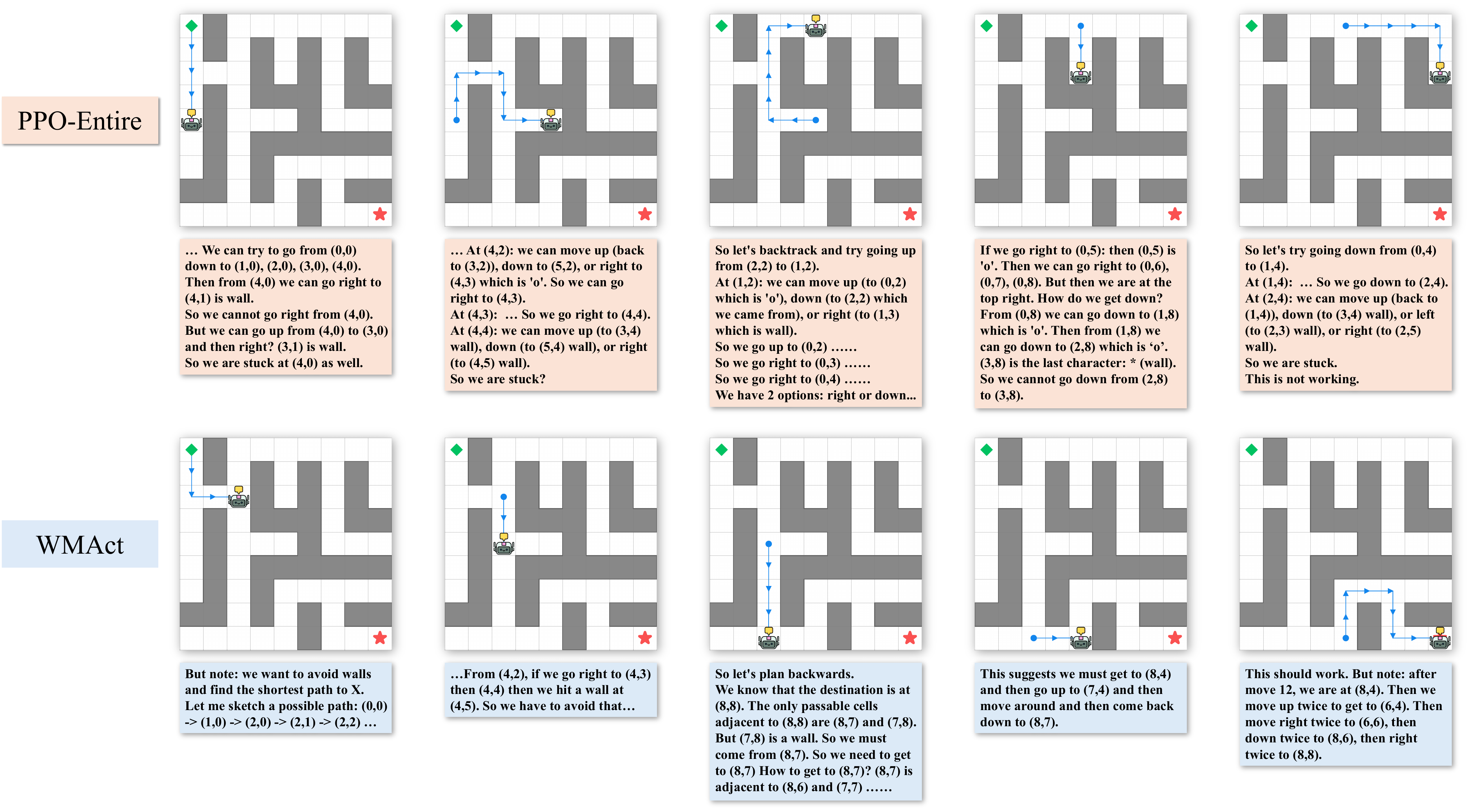}
    \caption{\textbf{Reasoning Trace Visualization: A qualitative comparison of PPO-Entire versus WMAct in a Maze task.} The top (PPO-Entire) demonstrates a more rigid, often enumeration-based or pre-computed planning approach, which frequently leads to suboptimal trajectories or impasses due to insufficient intermediate reflection and a lack of dynamic adaptation. In contrast, the bottom (WMAct) showcases the emergent interactive thinking and planning patterns of our model. Each step highlights WMAct's ability to analyze the current state, reflect on recent moves, and dynamically adjust its strategy, resulting in more robust self-correction mechanisms and efficient goal achievement.}
    \label{fig:ppo_entire_vs_wmact_maze_comparison}
\end{figure*}

\begin{figure*}[h!]
    \centering
    \includegraphics[width=1\linewidth]{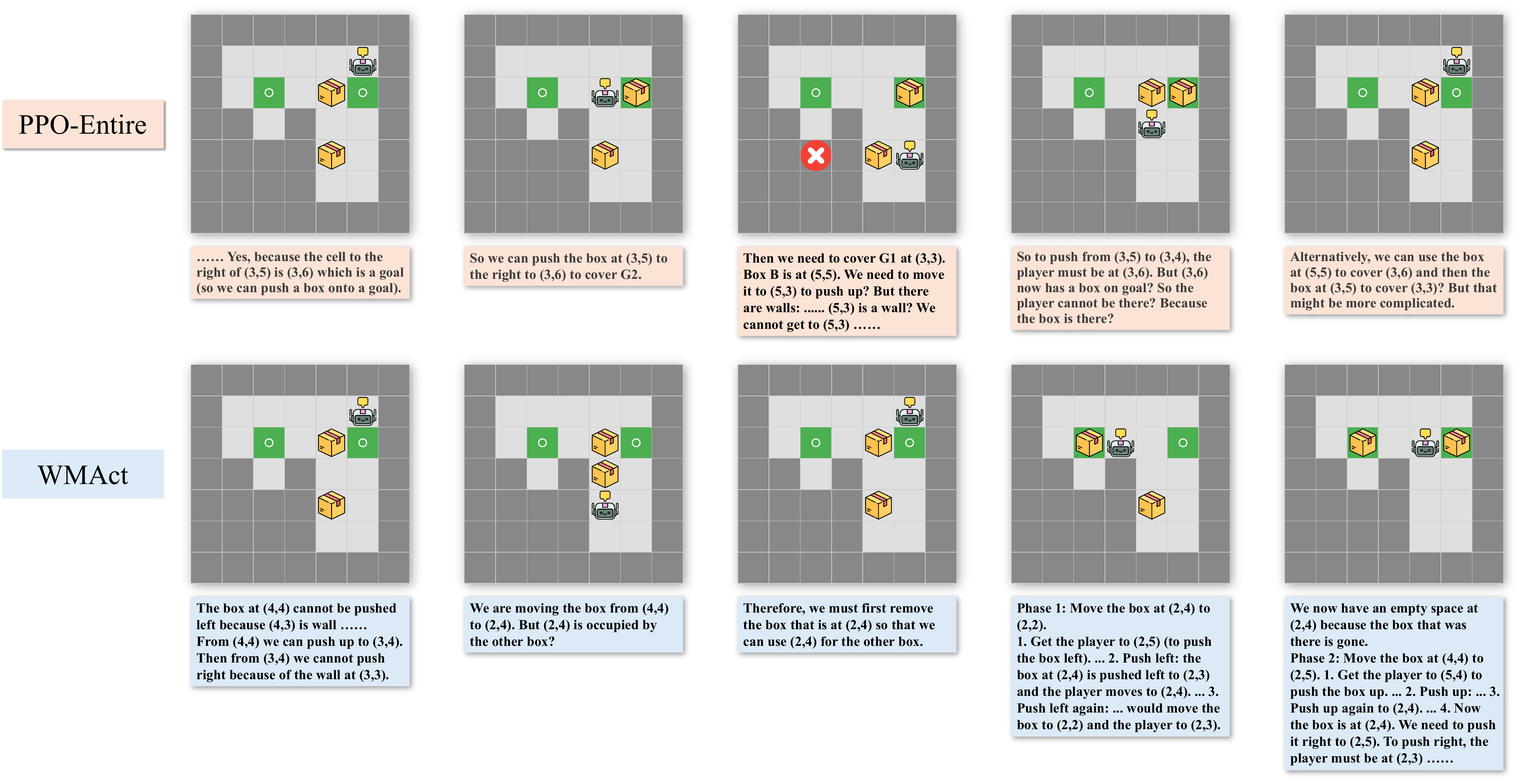}
    \caption{\textbf{Reasoning Trace Visualization: A qualitative comparison of PPO-Entire versus WMAct in a Sokoban task. }The top (PPO-Entire) demonstrates a common pitfall of model-based planning in Sokoban, where the agent, lacking robust foresight and intermediate reflection, gets trapped in an irreversible deadlock. In contrast, the bottom (WMAct) showcases the ability to systematically decompose the task and execute a non-greedy, long-term plan, demonstrating dynamic self-correction and resulting in an efficient, deadlock-free solution.}\label{fig:ppo_entire_vs_wmact_sokoban_comparison}
\end{figure*}

\begin{figure*}[h!]
    \centering
    \includegraphics[width=1\linewidth]{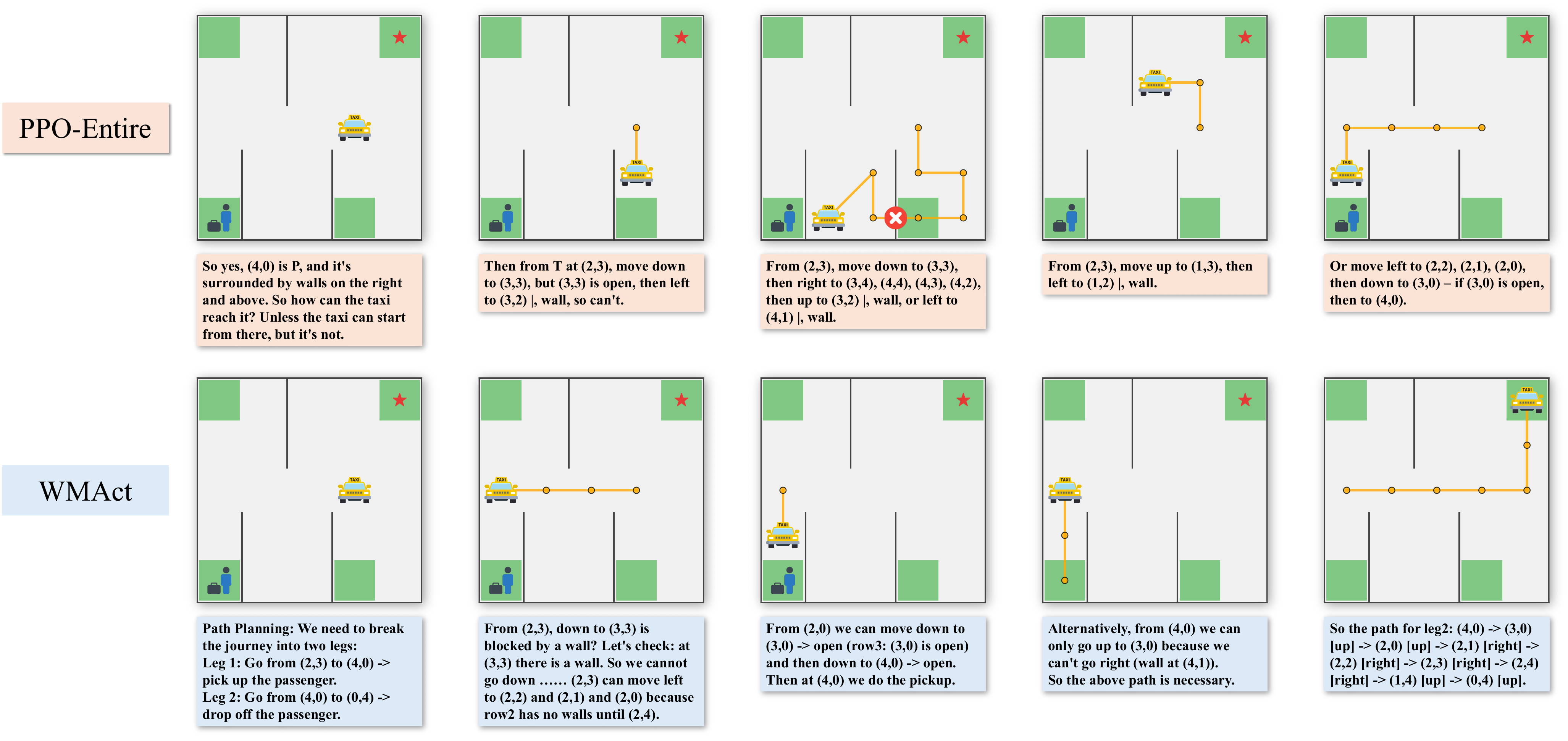}
    \caption{\textbf{Reasoning Trace Visualization: A qualitative comparison of PPO-Entire versus WMAct in a Taxi task.} The top (PPO-Entire) appears to lack robust global planning, often resorting to greedy local moves and struggling with dead-ends due to insufficient foresight regarding long-term path feasibility. In contrast, the bottom (WMAct) showcases superior global planning and sub-task decomposition, enabling it to systematically map out an efficient, wall-avoiding path.}
    \label{fig:ppo_entire_vs_wmact_taxi_comparison}
\end{figure*}

\section{Visual Case Studies and Reasoning Patterns}
\label{sec:case_studies}
To qualitatively assess the efficacy of World Model Reasoning, we present comparative reasoning trace visualizations across the Sokoban, Maze, and Taxi environments. These case studies collectively validate that WMAct's interactive thinking, driven by intermediate reflection, enables critical capabilities for solving complex, constraint-heavy tasks.

For the Maze (Figure \ref{fig:ppo_entire_vs_wmact_maze_comparison}), PPO-Entire adheres to a rigid, enumeration-based approach, frequently leading to suboptimal trajectories and impasses. Conversely, WMAct demonstrates superior long-term foresight, allowing it to quickly adapt to the complex topology and avoid unnecessary local detours, resulting in a significantly more efficient path.

For the Sokoban (Figure \ref{fig:ppo_entire_vs_wmact_sokoban_comparison}), PPO-Entire's greedy simulation leads the agent into an irreversible deadlock by failing to anticipate constraints. In contrast, WMAct showcases robust dynamic recovery and self-correction, swiftly identifying the flawed trajectory and executing a non-greedy, path-clearing maneuver to secure an efficient solution.

\newpage
For the Taxi (Figure \ref{fig:ppo_entire_vs_wmact_taxi_comparison}), PPO-Entire lacks the global view, often resorting to myopic local moves. WMAct, however, excels at systematic sub-task decomposition, enabling it to hierarchically structure the problem (e.g., pick up $\rightarrow$ drop off) and map out an optimal, wall-avoiding trajectory for the entire sequence. These results confirm that WMAct facilitates key capabilities—including robust recovery, environmental foresight, and hierarchical planning—that are essential for advanced sequential decision-making.

\clearpage
\section{Prompt Details}
\label{sec:prompt}
At the initial turn, the LLM receives a comprehensive prompt composed of three distinct, functional components: the \texttt{\{system prompt\}} for defining the agent's role; the \texttt{\{environment description\}} for providing rules, map legend, and current state; and the \texttt{\{action prompt\}} for strictly mandating the output format.
Crucially, the structural components of the prompt—the \texttt{\{system\ prompt\}} and the \texttt{\{action\ prompt\}}—are held strictly identical and un-engineered across all three distinct environments. This uniformity demonstrates the generality and minimal engineering overhead of our approach, ensuring that the framework's effectiveness relies on the explicit reasoning structure rather than domain-specific prompt tuning.

Consequently, the diverse and sophisticated thinking patterns evidenced in Section~\ref{sec:case_studies} are not hard-coded but emerge as learned capabilities, directly elicited by the model from the unique constraints and demands of each environment.
\begin{promptbox}{Sokoban Environment Description Example}
You are playing Sokoban, a puzzle game where you need to push boxes onto goal locations.\\
~\\~
Game state:
\begin{verbatim*}
#######
#.P..##
#GB.#.#
#..GG.#
#.BB..#
#..####
#######
\end{verbatim*}
~\\~
\textbf{Legend:}
\begin{itemize}
    \item P: Player
    \item B: Box
    \item G: Goal location
    \item *: Box on goal location
    \item \#: Wall
    \item .: Empty space
\end{itemize}
~\\~
Progress: 0/3 boxes on goals.\\
~\\~
\textbf{Available actions: }
\begin{itemize}
    \item push up, push down, push left, push right (to push a box if adjacent)
    \item move up, move down, move left, move right (to move without pushing)
    \item no operation (stay in place)
\end{itemize}
\end{promptbox}

\begin{promptbox}{System Prompt}
    You are a helpful assistant.
\end{promptbox}

\begin{promptbox}{Environment Action Prompt}
    For each turn, your response must follow a strict two-part structure. First, you must conduct your detailed reasoning within \textless{}think\textgreater{} and \textless{}/think\textgreater{} tags. Inside your thoughts, analyze the current state, reflect on your previous actions, and plan your next step. Second, immediately following your reasoning, you must provide your output within \textless{}action\textgreater{} and \textless{}/action\textgreater{} tags. The content of the \textless{}action\textgreater{} tag is critical and depends on the task. For interactive tasks, generate one or more commands separated by a semicolon, for example, \textless{}action\textgreater{}move up; move right\textless{}/action\textgreater{}. Your response for the turn concludes after you output the closing \textless{}/action\textgreater{} tag. Now, generate your thought and action based on the current state.
\end{promptbox}

\begin{promptbox}{Maze Environment Description Example}
You are an agent tasked with navigating a maze. Your goal is to move the player (P) to the destination (X) through a series of turns.
\\~\\
Maze Board:
\begin{verbatim}
X o o o o o o o o
* * o * o * * * *
o o o * o o o o o
o * * * * * o * *
o o o * o o o o o
o * * * * * o * o
o o o * o o o * o
o * o * o * o * *
o * o * o * o o P
\end{verbatim}
~\\~\\~
\textbf{Legend:}\\
`P' represents the player starting point\\
`X' represents the destination point\\
`o' represents empty space (passable)\\
`*' represents a wall (impassable)\\
~\\~\\
Your available moves are:\\
`move up': move player one cell up.\\
`move down': move player one cell down\\
`move left': move player one cell to the left\\
`move right': move player one cell to the right
\end{promptbox}

\begin{table*}[htbp] %
\begin{promptbox}{Taxi Environment Description Example}
You are an agent controlling a taxi in a 5x5 grid world. Your goal is to pick up the passenger and deliver them to the destination.
~\\~\\
Taxi Map:
\begin{verbatim}
+-----------------------------+
|  P  :     |     :     :  G  |
|     :     |     :     :     |
|     :     :     :     :     |
|     |     :     |  T  :     |
|  Y  |     :     |  D  :     |
+-----------------------------+
\end{verbatim}
`\textbar' represents a wall (an impassable barrier). \\
`:' represents an open path (a tile you can move through). \\
`T' represents the taxi position. \\
`P' represents the passenger position (if not in the taxi). \\
`D' represents the destination. \\
`R', 'G', 'Y', 'B' are special points. \\
`T/P'  means the taxi and the passenger are in the same cell, but the passenger has NOT been picked up yet. \\
`T(P)' means the taxi is carrying the passenger (the passenger has been picked up). \\
`T(P)/D' means the taxi carrying the passenger is at the destination. \\
`T/P/D' means the taxi and the passenger all arrive at the destination cell. \\
`P/D'  means the passenger and the destination are in the same cell (passenger not picked up). \\
~\\~
Your available actions are: \\
`move down': move the taxi one cell down. \\
`move up': move the taxi one cell up. \\ 
`move right': move the taxi one cell right. \\
`move left': move the taxi one cell left. \\
`pickup': pick up the passenger. This action is only valid when the taxi is at the passenger's location; otherwise, it will incur a penalty. \\
`dropoff': drop off the passenger. This action is only valid when the taxi is at the destination and carrying the passenger; otherwise, it will incur a penalty. \\
\end{promptbox}
\end{table*}


\begin{thebibliography}{53}
\providecommand{\natexlab}[1]{#1}
\providecommand{\url}[1]{\texttt{#1}}
\expandafter\ifx\csname urlstyle\endcsname\relax
  \providecommand{\doi}[1]{doi: #1}\else
  \providecommand{\doi}{doi: \begingroup \urlstyle{rm}\Url}\fi

\bibitem[Anthropic(2025)]{claude4p5}
Anthropic.
\newblock Introducing claude sonnet 4.5.
\newblock Blog post, 2025.
\newblock Accessed: September 2025.

\bibitem[{Art of Problem Solving}(2025)]{AoPS_AIME}
{Art of Problem Solving}.
\newblock {AIME Problems and Solutions}.
\newblock Official Website, 2025.
\newblock Accessed: September 2025.

\bibitem[{ByteDance-Seed}(2025)]{bytedance_seed_2025_beyondaime}
{ByteDance-Seed}.
\newblock {BeyondAIME: Advancing Math Reasoning Evaluation Beyond High School Olympiads}.
\newblock Hugging Face repository, 2025.
\newblock Accessed: November 2025.

\bibitem[Chen et~al.(2025)Chen, Gao, Liu, Huang, Sung, Zhou, Wu, and Chang]{chen2025g1}
Liang Chen, Hongcheng Gao, Tianyu Liu, Zhiqi Huang, Flood Sung, Xinyu Zhou, Yuxin Wu, and Baobao Chang.
\newblock G1: Bootstrapping perception and reasoning abilities of vision-language model via reinforcement learning.
\newblock \emph{arXiv preprint arXiv:2505.13426}, 2025.

\bibitem[Cheng et~al.(2025)Cheng, Hao, Liu, Zhou, Xie, Yao, Bian, Dey, Zhuang, Zha, Gu, Zhou, Wang, Li, Fan, She, Gao, Saparov, Killian, Li, Yurochkin, Xing, Liu, and Hu]{cheng2025revisiting}
Zhoujun Cheng, Shibo Hao, Tianyang Liu, Fan Zhou, Yutao Xie, Feng Yao, Yuexin Bian, Nilabjo Dey, Yonghao Zhuang, Yuheng Zha, Yi Gu, Kun Zhou, Yuqi Wang, Yuan Li, Richard Fan, Jianshu She, Chengqian Gao, Abulhair Saparov, Taylor~W. Killian, Haonan Li, Mikhail Yurochkin, Eric~P. Xing, Zhengzhong Liu, and Zhiting Hu.
\newblock Revisiting reinforcement learning for {LLM} reasoning from a cross-domain perspective.
\newblock In \emph{The Thirty-ninth Annual Conference on Neural Information Processing Systems Datasets and Benchmarks Track}, 2025.

\bibitem[Comanici et~al.(2025)Comanici, Bieber, Schaekermann, Pasupat, Sachdeva, Dhillon, Blistein, Ram, Zhang, Rosen, et~al.]{comanici2025gemini}
Gheorghe Comanici, Eric Bieber, Mike Schaekermann, Ice Pasupat, Noveen Sachdeva, Inderjit Dhillon, Marcel Blistein, Ori Ram, Dan Zhang, Evan Rosen, et~al.
\newblock Gemini 2.5: Pushing the frontier with advanced reasoning, multimodality, long context, and next generation agentic capabilities.
\newblock \emph{arXiv preprint arXiv:2507.06261}, 2025.

\bibitem[Guo et~al.(2025{\natexlab{a}})Guo, Wu, Zhu, Leng, Shi, Chen, Fan, Wang, Jiang, Wang, et~al.]{guo2025seed1}
Dong Guo, Faming Wu, Feida Zhu, Fuxing Leng, Guang Shi, Haobin Chen, Haoqi Fan, Jian Wang, Jianyu Jiang, Jiawei Wang, et~al.
\newblock Seed1. 5-vl technical report.
\newblock \emph{arXiv preprint arXiv:2505.07062}, 2025{\natexlab{a}}.

\bibitem[Guo et~al.(2025{\natexlab{b}})Guo, Yang, Zhang, Song, Wang, Zhu, Xu, Zhang, Ma, Bi, Zhang, Yu, Wu, Wu, Gou, Shao, Li, Gao, Liu, Xue, and Zhang]{guo2025deepseek}
D. Guo, D. Yang, H. Zhang, J. Song, P. Wang, Q. Zhu, R. Xu, R. Zhang, S. Ma, X. Bi, X. Zhang, X. Yu, Y. Wu, Z.~F. Wu, Z. Gou, Z. Shao, Z. Li, Z. Gao, A. Liu, B. Xue, and Z. Zhang.
\newblock {DeepSeek-R1 incentivizes reasoning in LLMs through reinforcement learning}.
\newblock \emph{Nature}, 645\penalty0 (8081):\penalty0 633--638, 2025{\natexlab{b}}.

\bibitem[He et~al.(2025)He, Liu, Liu, Yan, Wang, Cheng, Zhang, Zhang, Xu, Shen, et~al.]{skywork-or1-2025}
Jujie He, Jiacai Liu, Chris~Yuhao Liu, Rui Yan, Chaojie Wang, Peng Cheng, Xiaoyu Zhang, Fuxiang Zhang, Jiacheng Xu, Wei Shen, et~al.
\newblock Skywork open reasoner 1 technical report.
\newblock \emph{arXiv preprint arXiv:2505.22312}, 2025.

\bibitem[{HMMT Staff}(2025)]{HMMT2025}
{HMMT Staff}.
\newblock {HMMT 2025 Problems and Solutions}.
\newblock Official Website, 2025.
\newblock Accessed: November 2025.

\bibitem[Hong et~al.(2025)Hong, Yu, Gu, Wang, Gan, Tang, Cheng, Qi, Ji, Pan, et~al.]{hong2025glm}
Wenyi Hong, Wenmeng Yu, Xiaotao Gu, Guo Wang, Guobing Gan, Haomiao Tang, Jiale Cheng, Ji Qi, Junhui Ji, Lihang Pan, et~al.
\newblock Glm-4.1 v-thinking: Towards versatile multimodal reasoning with scalable reinforcement learning.
\newblock \emph{arXiv preprint arXiv:2507.01006}, 2025.

\bibitem[Hu et~al.(2025)Hu, Zhang, Han, Jiang, Zhang, and Shum]{hu2025open}
Jingcheng Hu, Yinmin Zhang, Qi Han, Daxin Jiang, Xiangyu Zhang, and Heung-Yeung Shum.
\newblock Open-reasoner-zero: An open source approach to scaling up reinforcement learning on the base model.
\newblock \emph{arXiv preprint arXiv:2503.24290}, 2025.

\bibitem[Hurst et~al.(2024)Hurst, Lerer, Goucher, Perelman, Ramesh, Clark, Ostrow, Welihinda, Hayes, Radford, et~al.]{hurst2024gpt}
Aaron Hurst, Adam Lerer, Adam~P Goucher, Adam Perelman, Aditya Ramesh, Aidan Clark, AJ Ostrow, Akila Welihinda, Alan Hayes, Alec Radford, et~al.
\newblock Gpt-4o system card.
\newblock \emph{arXiv preprint arXiv:2410.21276}, 2024.

\bibitem[Jaech et~al.(2024)Jaech, Kalai, Lerer, Richardson, El-Kishky, Low, Helyar, Madry, Beutel, Carney, et~al.]{jaech2024openai}
Aaron Jaech, Adam Kalai, Adam Lerer, Adam Richardson, Ahmed El-Kishky, Aiden Low, Alec Helyar, Aleksander Madry, Alex Beutel, Alex Carney, et~al.
\newblock Openai o1 system card.
\newblock \emph{arXiv preprint arXiv:2412.16720}, 2024.

\bibitem[Jain et~al.(2025)Jain, Han, Gu, Li, Yan, Zhang, Wang, Solar-Lezama, Sen, and Stoica]{jain2025livecodebench}
Naman Jain, King Han, Alex Gu, Wen-Ding Li, Fanjia Yan, Tianjun Zhang, Sida Wang, Armando Solar-Lezama, Koushik Sen, and Ion Stoica.
\newblock Livecodebench: Holistic and contamination free evaluation of large language models for code.
\newblock In \emph{The Thirteenth International Conference on Learning Representations}, 2025.

\bibitem[Liao et~al.(2025)Liao, Gu, Sui, Zhu, Lu, Tang, Sun, and Yang]{liao2025think}
Yi Liao, Yu Gu, Yuan Sui, Zining Zhu, Yifan Lu, Guohua Tang, Zhongqian Sun, and Wei Yang.
\newblock Think in games: Learning to reason in games via reinforcement learning with large language models.
\newblock \emph{arXiv preprint arXiv:2508.21365}, 2025.

\bibitem[Liu et~al.(2025)Liu, Xie, Ding, Li, Yang, Wu, Wang, Sun, Liu, Wang, Ye, Li, Dong, Yu, Lu, Mo, Yan, Tian, Zhang, Huang, Liu, Su, Luo, Yue, Qi, Chen, Zhou, Qiao, Chen, and Wang]{liu2025scalecua}
Zhaoyang Liu, Jingjing Xie, Zichen Ding, Zehao Li, Bowen Yang, Zhenyu Wu, Xuehui Wang, Qiushi Sun, Shi Liu, Weiyun Wang, Shenglong Ye, Qingyun Li, Xuan Dong, Yue Yu, Chenyu Lu, YunXiang Mo, Yao Yan, Zeyue Tian, Xiao Zhang, Yuan Huang, Yiqian Liu, Weijie Su, Gen Luo, Xiangyu Yue, Biqing Qi, Kai Chen, Bowen Zhou, Yu Qiao, Qifeng Chen, and Wenhai Wang.
\newblock Scalecua: Scaling open-source computer use agents with cross-platform data.
\newblock \emph{arXiv preprint arXiv:2509.15221}, 2025.

\bibitem[OpenAI(2025{\natexlab{a}})]{openaigpt5}
OpenAI.
\newblock Gpt-5 system card.
\newblock Blog post, 2025{\natexlab{a}}.
\newblock Accessed: August 2025.

\bibitem[OpenAI(2025{\natexlab{b}})]{openaio1}
OpenAI.
\newblock Openai o1 system card.
\newblock Blog post, 2025{\natexlab{b}}.
\newblock Accessed: December 2025.

\bibitem[OpenAI(2025{\natexlab{c}})]{openaio3}
OpenAI.
\newblock Openai o3 and o4-mini system card.
\newblock Blog post, 2025{\natexlab{c}}.
\newblock Accessed: April 2025.

\bibitem[Pourreza et~al.(2025)Pourreza, Talaei, Sun, Wan, Li, Mirhoseini, Saberi, and Arik]{pourreza2025reasoningsql}
Mohammadreza Pourreza, Shayan Talaei, Ruoxi Sun, Xingchen Wan, Hailong Li, Azalia Mirhoseini, Amin Saberi, and Sercan~O Arik.
\newblock Reasoning-{SQL}: Reinforcement learning with {SQL} tailored partial rewards for reasoning-enhanced text-to-{SQL}.
\newblock In \emph{The Second Conference on Language Modeling}, 2025.

\bibitem[Rein et~al.(2024)Rein, Hou, Stickland, Petty, Pang, Dirani, Michael, and Bowman]{rein2024gpqa}
David Rein, Betty~Li Hou, Asa~Cooper Stickland, Jackson Petty, Richard~Yuanzhe Pang, Julien Dirani, Julian Michael, and Samuel~R. Bowman.
\newblock {GPQA}: A graduate-level google-proof q\&a benchmark.
\newblock In \emph{First Conference on Language Modeling}, 2024.

\bibitem[Schulman et~al.(2017)Schulman, Wolski, Dhariwal, Radford, and Klimov]{schulman2017proximal}
John Schulman, Filip Wolski, Prafulla Dhariwal, Alec Radford, and Oleg Klimov.
\newblock Proximal policy optimization algorithms.
\newblock \emph{arXiv preprint arXiv:1707.06347}, 2017.

\bibitem[Shang et~al.(2025)Shang, Liu, Zhu, Zhang, Xu, Guan, Zhang, Dong, Zhou, Zhang, et~al.]{shang2025rstar2}
Ning Shang, Yifei Liu, Yi Zhu, Li~Lyna Zhang, Weijiang Xu, Xinyu Guan, Buze Zhang, Bingcheng Dong, Xudong Zhou, Bowen Zhang, et~al.
\newblock rstar2-agent: Agentic reasoning technical report.
\newblock \emph{arXiv preprint arXiv:2508.20722}, 2025.

\bibitem[Silver et~al.(2017)Silver, Schrittwieser, Simonyan, Antonoglou, Huang, Guez, Hubert, baker, Lai, Bolton, Chen, Lillicrap, Hui, Sifre, van~den Driessche, Graepel, and Hassabis]{Silver2017MasteringTG}
David Silver, Julian Schrittwieser, Karen Simonyan, Ioannis Antonoglou, Aja Huang, Arthur Guez, Thomas Hubert, Lucas baker, Matthew Lai, Adrian Bolton, Yutian Chen, Timothy~P. Lillicrap, Fan Hui, L. Sifre, George van~den Driessche, Thore Graepel, and Demis Hassabis.
\newblock Mastering the game of go without human knowledge.
\newblock \emph{Nature}, 550:\penalty0 354--359, 2017.

\bibitem[Silver et~al.(2018)Silver, Hubert, Schrittwieser, Antonoglou, Lai, Guez, Lanctot, Sifre, Kumaran, Graepel, et~al.]{silver2017mastering}
David Silver, Thomas Hubert, Julian Schrittwieser, Ioannis Antonoglou, Matthew Lai, Arthur Guez, Marc Lanctot, Laurent Sifre, Dharshan Kumaran, Thore Graepel, et~al.
\newblock {A general reinforcement learning algorithm that masters chess, shogi, and Go through self-play}.
\newblock \emph{Science}, 362\penalty0 (6419):\penalty0 1140--1144, 2018.

\bibitem[Sprague et~al.(2025)Sprague, Yin, Rodriguez, Jiang, Wadhwa, Singhal, Zhao, Ye, Mahowald, and Durrett]{sprague2024cot}
Zayne~Rea Sprague, Fangcong Yin, Juan~Diego Rodriguez, Dongwei Jiang, Manya Wadhwa, Prasann Singhal, Xinyu Zhao, Xi Ye, Kyle Mahowald, and Greg Durrett.
\newblock To cot or not to cot? chain-of-thought helps mainly on math and symbolic reasoning.
\newblock In \emph{The Thirteenth International Conference on Learning Representations}, 2025.

\bibitem[Su et~al.(2025)Su, Zhang, Li, Chen, Wang, Song, Wang, Li, Wu, Chen, et~al.]{su2025scaling}
Liangcai Su, Zhen Zhang, Guangyu Li, Zhuo Chen, Chenxi Wang, Maojia Song, Xinyu Wang, Kuan Li, Jialong Wu, Xuanzhong Chen, et~al.
\newblock Scaling agents via continual pre-training.
\newblock \emph{arXiv preprint arXiv:2509.13310}, 2025.

\bibitem[Sutton(2019)]{Sutton2019BitterLesson}
Richard Sutton.
\newblock The bitter lesson.
\newblock Blog post, 2019.
\newblock Accessed: August 2024.

\bibitem[Sutton and Barto(1998)]{712192}
R.S. Sutton and A.G. Barto.
\newblock Reinforcement learning: An introduction.
\newblock \emph{IEEE Transactions on Neural Networks}, 9\penalty0 (5):\penalty0 1054--1054, 1998.

\bibitem[Sutton et~al.(1999)Sutton, McAllester, Singh, and Mansour]{NIPS1999_464d828b}
Richard~S Sutton, David McAllester, Satinder Singh, and Yishay Mansour.
\newblock Policy gradient methods for reinforcement learning with function approximation.
\newblock In \emph{Advances in Neural Information Processing Systems}. MIT Press, 1999.

\bibitem[Team et~al.(2025)Team, Du, Gao, Xing, Jiang, Chen, Li, Xiao, Du, Liao, et~al.]{team2025kimi}
Kimi Team, Angang Du, Bofei Gao, Bowei Xing, Changjiu Jiang, Cheng Chen, Cheng Li, Chenjun Xiao, Chenzhuang Du, Chonghua Liao, et~al.
\newblock Kimi k1. 5: Scaling reinforcement learning with llms.
\newblock \emph{arXiv preprint arXiv:2501.12599}, 2025.

\bibitem[Team(2025)]{qwen3technicalreport}
Qwen Team.
\newblock Qwen3 technical report, 2025.

\bibitem[Wang et~al.(2025{\natexlab{a}})Wang, Wang, Wan, Huang, Hu, Jia, Nie, Li, Chen, Chen, et~al.]{wang2025step}
Bin Wang, Bojun Wang, Changyi Wan, Guanzhe Huang, Hanpeng Hu, Haonan Jia, Hao Nie, Mingliang Li, Nuo Chen, Siyu Chen, et~al.
\newblock Step-3 is large yet affordable: Model-system co-design for cost-effective decoding.
\newblock \emph{arXiv preprint arXiv:2507.19427}, 2025{\natexlab{a}}.

\bibitem[Wang et~al.(2025{\natexlab{b}})Wang, Zhang, Wang, Gao, Li, Wang, Chen, Lu, Yang, Wang, Krishna, Wu, Fei-Fei, Choi, and Li]{wang2025vagen}
Kangrui Wang, Pingyue Zhang, Zihan Wang, Yaning Gao, Linjie Li, Qineng Wang, Hanyang Chen, Yiping Lu, Zhengyuan Yang, Lijuan Wang, Ranjay Krishna, Jiajun Wu, Li Fei-Fei, Yejin Choi, and Manling Li.
\newblock {VAGEN}: Reinforcing world model reasoning for multi-turn {VLM} agents.
\newblock In \emph{The Thirty-ninth Annual Conference on Neural Information Processing Systems}, 2025{\natexlab{b}}.

\bibitem[Wang et~al.(2025{\natexlab{c}})Wang, Gao, Gu, Pu, Cui, Wei, Liu, Jing, Ye, Shao, et~al.]{wang2025internvl3}
Weiyun Wang, Zhangwei Gao, Lixin Gu, Hengjun Pu, Long Cui, Xingguang Wei, Zhaoyang Liu, Linglin Jing, Shenglong Ye, Jie Shao, et~al.
\newblock Internvl3.5: Advancing open-source multimodal models in versatility, reasoning, and efficiency.
\newblock \emph{arXiv preprint arXiv:2508.18265}, 2025{\natexlab{c}}.

\bibitem[Wang et~al.(2024)Wang, Ma, Zhang, Ni, Chandra, Guo, Ren, Arulraj, He, Jiang, Li, Ku, Wang, Zhuang, Fan, Yue, and Chen]{wang2024mmlupro}
Yubo Wang, Xueguang Ma, Ge Zhang, Yuansheng Ni, Abhranil Chandra, Shiguang Guo, Weiming Ren, Aaran Arulraj, Xuan He, Ziyan Jiang, Tianle Li, Max Ku, Kai Wang, Alex Zhuang, Rongqi Fan, Xiang Yue, and Wenhu Chen.
\newblock {MMLU}-pro: A more robust and challenging multi-task language understanding benchmark.
\newblock In \emph{The Thirty-eight Conference on Neural Information Processing Systems Datasets and Benchmarks Track}, 2024.

\bibitem[Wang et~al.(2025{\natexlab{d}})Wang, Wang, Wang, Zhang, Li, Yang, Jin, Yu, Nguyen, Liu, et~al.]{wang2025ragen}
Zihan Wang, Kangrui Wang, Qineng Wang, Pingyue Zhang, Linjie Li, Zhengyuan Yang, Xing Jin, Kefan Yu, Minh~Nhat Nguyen, Licheng Liu, et~al.
\newblock Ragen: Understanding self-evolution in llm agents via multi-turn reinforcement learning.
\newblock \emph{arXiv preprint arXiv:2504.20073}, 2025{\natexlab{d}}.

\bibitem[Wei et~al.(2024)Wei, Yin, Li, Wang, Zhao, Sun, Ge, Zhang, and Jiang]{wei2024slow}
Haoran Wei, Youyang Yin, Yumeng Li, Jia Wang, Liang Zhao, Jianjian Sun, Zheng Ge, Xiangyu Zhang, and Daxin Jiang.
\newblock Slow perception: Let's perceive geometric figures step-by-step.
\newblock \emph{arXiv preprint arXiv:2412.20631}, 2024.

\bibitem[Wei et~al.(2025{\natexlab{a}})Wei, Duchenne, Copet, Carbonneaux, Zhang, Fried, Synnaeve, Singh, and Wang]{wei2025swe}
Yuxiang Wei, Olivier Duchenne, Jade Copet, Quentin Carbonneaux, Lingming Zhang, Daniel Fried, Gabriel Synnaeve, Rishabh Singh, and Sida Wang.
\newblock {SWE}-{RL}: Advancing {LLM} reasoning via reinforcement learning on open software evolution.
\newblock In \emph{The Thirty-ninth Annual Conference on Neural Information Processing Systems}, 2025{\natexlab{a}}.

\bibitem[Wei et~al.(2025{\natexlab{b}})Wei, Zhao, Sun, Lin, jisheng yin, Hu, Zhang, Yu, Lv, Weng, Wang, Han, Ge, Zhang, Jiang, and Patel]{wei2025open}
Yana Wei, Liang Zhao, Jianjian Sun, Kangheng Lin, jisheng yin, Jingcheng Hu, Yinmin Zhang, En Yu, Haoran Lv, Zejia Weng, Jia Wang, Qi Han, Zheng Ge, Xiangyu Zhang, Daxin Jiang, and Vishal~M. Patel.
\newblock Open vision reasoner: Transferring linguistic cognitive behavior for visual reasoning.
\newblock In \emph{The Thirty-ninth Annual Conference on Neural Information Processing Systems}, 2025{\natexlab{b}}.

\bibitem[White et~al.(2024)White, Dooley, Roberts, Pal, Feuer, Jain, Shwartz-Ziv, Jain, Saifullah, Naidu, et~al.]{white2024livebench}
Colin White, Samuel Dooley, Manley Roberts, Arka Pal, Ben Feuer, Siddhartha Jain, Ravid Shwartz-Ziv, Neel Jain, Khalid Saifullah, Siddartha Naidu, et~al.
\newblock Livebench: A challenging, contamination-free llm benchmark.
\newblock \emph{arXiv preprint arXiv:2406.19314}, 4, 2024.

\bibitem[Xie et~al.(2025)Xie, Gao, Ren, Luo, Hong, Dai, Zhou, Qiu, Wu, and Luo]{xie2025logic}
Tian Xie, Zitian Gao, Qingnan Ren, Haoming Luo, Yuqian Hong, Bryan Dai, Joey Zhou, Kai Qiu, Zhirong Wu, and Chong Luo.
\newblock Logic-rl: Unleashing llm reasoning with rule-based reinforcement learning.
\newblock \emph{arXiv preprint arXiv:2502.14768}, 2025.

\bibitem[Xue et~al.(2025)Xue, Zheng, Liu, Li, Zheng, Ma, and An]{xue2025simpletir}
Zhenghai Xue, Longtao Zheng, Qian Liu, Yingru Li, Xiaosen Zheng, Zejun Ma, and Bo An.
\newblock Simpletir: End-to-end reinforcement learning for multi-turn tool-integrated reasoning.
\newblock \emph{arXiv preprint arXiv:2509.02479}, 2025.

\bibitem[Yang et~al.(2024)Yang, Yang, Zhang, Hui, Zheng, Yu, Li, Liu, Huang, Wei, et~al.]{yang2024qwen2.5}
An Yang, Baosong Yang, Beichen Zhang, Binyuan Hui, Bo Zheng, Bowen Yu, Chengyuan Li, Dayiheng Liu, Fei Huang, Haoran Wei, et~al.
\newblock Qwen2. 5 technical report.
\newblock \emph{arXiv preprint arXiv:2412.15115}, 2024.

\bibitem[Yu et~al.(2025{\natexlab{a}})Yu, Lin, Zhao, jisheng yin, Wei, Peng, Wei, Sun, Han, Ge, Zhang, Jiang, Wang, and Tao]{yu2025perceptionr}
En Yu, Kangheng Lin, Liang Zhao, jisheng yin, Yana Wei, Yuang Peng, Haoran Wei, Jianjian Sun, Chunrui Han, Zheng Ge, Xiangyu Zhang, Daxin Jiang, Jingyu Wang, and Wenbing Tao.
\newblock Perception-r1: Pioneering perception policy with reinforcement learning.
\newblock In \emph{The Thirty-ninth Annual Conference on Neural Information Processing Systems}, 2025{\natexlab{a}}.

\bibitem[Yu et~al.(2025{\natexlab{b}})Yu, Lin, Zhao, Wei, Zhu, Wei, Sun, Ge, Zhang, Wang, and Tao]{yu2025unhackable}
En Yu, Kangheng Lin, Liang Zhao, Yana Wei, Zining Zhu, Haoran Wei, Jianjian Sun, Zheng Ge, Xiangyu Zhang, Jingyu Wang, and Wenbing Tao.
\newblock Unhackable temporal reward for scalable video {MLLM}s.
\newblock In \emph{The Thirteenth International Conference on Learning Representations}, 2025{\natexlab{b}}.

\bibitem[Yu et~al.(2025{\natexlab{c}})Yu, Zhang, Zhu, Yuan, Zuo, YuYue, Dai, Fan, Liu, Liu, Liu, Liu, Lin, Lin, Ma, Sheng, Tong, Zhang, Zhang, Zhang, Zhang, Zhu, Zhu, Chen, Chen, Wang, Yu, Song, Wei, Zhou, Liu, Ma, Zhang, Yan, Wu, and Wang]{yu2025dapo}
Qiying Yu, Zheng Zhang, Ruofei Zhu, Yufeng Yuan, Xiaochen Zuo, YuYue, Weinan Dai, Tiantian Fan, Gaohong Liu, Juncai Liu, LingJun Liu, Xin Liu, Haibin Lin, Zhiqi Lin, Bole Ma, Guangming Sheng, Yuxuan Tong, Chi Zhang, Mofan Zhang, Ru Zhang, Wang Zhang, Hang Zhu, Jinhua Zhu, Jiaze Chen, Jiangjie Chen, Chengyi Wang, Hongli Yu, Yuxuan Song, Xiangpeng Wei, Hao Zhou, Jingjing Liu, Wei-Ying Ma, Ya-Qin Zhang, Lin Yan, Yonghui Wu, and Mingxuan Wang.
\newblock {DAPO}: An open-source {LLM} reinforcement learning system at scale.
\newblock In \emph{The Thirty-ninth Annual Conference on Neural Information Processing Systems}, 2025{\natexlab{c}}.

\bibitem[Zeng et~al.(2025)Zeng, Huang, Liu, Liu, He, Ma, and He]{zeng2025simplerlzooinvestigatingtamingzero}
Weihao Zeng, Yuzhen Huang, Qian Liu, Wei Liu, Keqing He, Zejun Ma, and Junxian He.
\newblock Simplerl-zoo: Investigating and taming zero reinforcement learning for open base models in the wild.
\newblock In \emph{The Second Conference on Language Modeling}, 2025.

\bibitem[Zhang et~al.(2025{\natexlab{a}})Zhang, Chen, Liu, Xue, Liao, Liu, Wang, Ning, Chen, Fu, et~al.]{zhang2025agent}
Kai Zhang, Xiangchao Chen, Bo Liu, Tianci Xue, Zeyi Liao, Zhihan Liu, Xiyao Wang, Yuting Ning, Zhaorun Chen, Xiaohan Fu, et~al.
\newblock Agent learning via early experience.
\newblock \emph{arXiv preprint arXiv:2510.08558}, 2025{\natexlab{a}}.

\bibitem[Zhang et~al.(2025{\natexlab{b}})Zhang, Chen, Li, Tu, and Li]{zhang2025rlvmr}
Zijing Zhang, Ziyang Chen, Mingxiao Li, Zhaopeng Tu, and Xiaolong Li.
\newblock Rlvmr: Reinforcement learning with verifiable meta-reasoning rewards for robust long-horizon agents.
\newblock \emph{arXiv preprint arXiv:2507.22844}, 2025{\natexlab{b}}.

\bibitem[Zhou et~al.(2024)Zhou, Zanette, Pan, Levine, and Kumar]{zhou2024archer}
Yifei Zhou, Andrea Zanette, Jiayi Pan, Sergey Levine, and Aviral Kumar.
\newblock Ar{CH}er: Training language model agents via hierarchical multi-turn {RL}.
\newblock In \emph{Forty-first International Conference on Machine Learning}, 2024.

\bibitem[Zhu et~al.(2025)Zhu, Zhao, Lin, Yang, Yu, Liu, Wei, Sun, Ge, and Zhang]{zhu2025perpo}
Zining Zhu, Liang Zhao, Kangheng Lin, Jinze Yang, En Yu, Chenglong Liu, Haoran Wei, Jianjian Sun, Zheng Ge, and Xiangyu Zhang.
\newblock Perpo: Perceptual preference optimization via discriminative rewarding.
\newblock \emph{arXiv preprint arXiv:2502.04371}, 2025.

\end{thebibliography}
\end{document}